\documentclass[twoside,11pt]{article}
\usepackage{natbib}
\usepackage{amsmath,amsfonts}
\usepackage{longtable}
\usepackage{graphicx}
\usepackage{url}
\usepackage{multirow}

\usepackage{caption}
\usepackage{subcaption}
\newcommand{\error}{\epsilon}
\newcommand{\jhat}{\hat{\jmath}}

\begin{document}

%\begin{frontmatter}

%\jmlrheading{}{2016}{}{}{}{Jacques Wainer} 
%\ShortHeadings{Comparison of 14 different families of classification  algorithms}{Wainer}

%\firstpageno{1}

\title{Comparison of 14 different families of classification
  algorithms on 115
  binary datasets}

\author{ Jacques Wainer \\email:  wainer@ic.unicamp.br\\
 Computing Institute\\
University of Campinas\\
Campinas, SP, 13083-852, Brazil}

%\editor{}

\maketitle

\begin{abstract}
We tested 14 very different classification algorithms (random forest,
gradient boosting machines, SVM - linear, polynomial, and RBF
- 1-hidden-layer  neural
nets, extreme learning machines, k-nearest neighbors and a bagging of
knn, naive Bayes, learning vector quantization, elastic net logistic
regression, sparse linear discriminant analysis, and a boosting of
linear classifiers) on 115 real life binary datasets. We followed the Demsar
analysis and found that the three best classifiers (random forest,
gbm and RBF SVM) are not significantly different from each other. We
also discuss that a change of less then 0.0112 in the error rate should be
considered as an irrelevant change, and used a Bayesian ANOVA analysis
to conclude that with high probability the differences between these
three classifiers is not of practical consequence. We also verified
the execution time of ``standard implementations'' of these algorithms
and concluded that RBF SVM is the fastest (significantly so) both in
training time and in training plus testing time.

%\begin{keywords}
\textbf{keywords:} Classification algorithms; comparison; binary problems; Demsar
procedure; Bayesian analysis
%\end{keywords}

\end{abstract}

\section{Introduction}

\cite{delgado14} evaluated 179 different implementations of
classification algorithms (from 17 different families of algoritms) on
121 public datasets. We believe that such highly empirical research
are very important both for researchers in machine learning and
specially for practitioners. For researchers, this form of research
allows them to focus their efforts on more likely useful
endeavors. For example, if a researcher is interested in developing
algorithms for very large classification problems, it is probably more
useful to develop a big-data random forest (which is the family of
classification algoritms with best performance according to
\cite{delgado14}) than to do it for Bayesian networks (mainly naive
Bayes) or even Nearest Neighbors methods, which perform worse than
random forests.

For practitioners, this form of research is even more important. 
Practitioners in machine learning will have limited resources, time,
and expertise to test many different classification algorithms on
their problem, and this form of research will allow them to focus on
the most likely useful algorithms. 

Despite its importance and breath, we believe that \cite{delgado14}
had some ``imperfections'' which we address in this
research. The ``imperfections'' are discussed below as the extensions
we carried in this paper:
\begin{itemize}
\item We used the same set of datasets, but we transform them so that
  the problems are all binary. Many classification algorithms are
  defined only to binary problems, for example the SVM family, logistic
  regression, among others. Of course there are meta-extensions of such
  algorithms to multi-class problems, for example, one-vs-one,
  one-vs-all, error correcting output coding (ECOC)
  \citep{dietterich1995solving}, stacked generalization
  \citep{wolpert1992stacked}, pairwise coupling
  \citep{hastie1998classification}, among others.  Also, there are
  alternative formulations for specific binary classifiers to deal with
  multiclass, for example, \cite{franc2002multi} for SVM, \cite{engel1988polytomous} for
  logistic regression, and so on. 

For the algorithms that are intrinsically binary, the application to
multiclass problems poses two problems. The first is that one has to
decide on which meta-extension to use, or if one should use the
specific multiclass formulation of the algorithm. In some cases, one
meta-extension is implemented by default, and the user must be aware
that this decision was already made for him/her. For example, the
libSVM default approach is one-vs-one. But a second, more subtle  problem is the
search for hyperparameters: it is very common that each combination of
hyperparameters are tested only once for all classifiers in the
one-vs-one solution. That is, all $n (n-1)/2$ classifiers in a
one-vs-one solution has the same set of hyperparameters, and that may
cause a decrease in accuracy in comparison to the case in which
classifier is allow to choose its one set of hyperparameter. Thus, on multiclass
problems, those intrinsically binary algorithms could be facing many
disadvantages. 

On the issue of binary classifiers, \cite{delgado14} did not include
in their comparisons the gradient boosting machine (\texttt{gbm}) algorithm,
considered a very competitive algorithm for classification problems because,
as reported in  \cite{delgado-gbm},  the implementation did not work in
multiclass problems. We included \texttt{gbm} in our comparison.

\item We reduced the number of classifiers to only a few classes/algorithms and
  not different implementations of the same
  algorithm. \cite{delgado14} compared an impressing 179 different
  classification programs, but it was unclear how many were just
  different implementations of the same algorithm, and how many were
  variations within the same ``family'' of algorithms. We believe that for
  practitioner and  research communities, it is more useful to
  have an understanding of how different families of algorithms rank
  in relation to each other. For the practitioner, which should have
  more limited access to the different algorithms, this knowledge
  allow them to order which algorithms should be applied first to
  their particular problem. 

In fact, \cite{delgado14}  also perform an analysis of their results
based on the algorithm's  ``family'', but they have difficulty of
extracting useful information from this analysis, since in most
cases, different ``implementations'' in the same family have widely
different results. In one analysis, they rank the families by the
worse performing member, which does not provide useful
information. But in the end, their conclusions are mainly centered on
the families of classifiers, because that is the most useful level of
knowledge. From the abstract of the paper:
\begin{quote}
  A few models are clearly better than the remaining ones: random
  forest, SVM with Gaussian and polynomial kernels, extreme learning
  machine with Gaussian kernel, C5.0 and avNNet (a committee of
  multi-layer perceptrons implemented in R with the caret
  package). The random forest is clearly the best family of
  classifiers (3 out of 5 bests classifiers are RF), followed by SVM
  (4 classifiers in the top-10), neural networks and boosting
  ensembles (5 and 3 members in the top-20, respectively).
\end{quote}

\item We performed a more careful search for hyperparameters for each
  classification algorithm. Given \cite{delgado14} daunting task of
  testing 179 programs, they had to rely on default values for the
  hyperparameters which may lead to suboptimal results. Since we are
  working with significantly fewer algorithms, we could spend more time
  selecting reasonable ranges of values for the hyperparameters. In
  particular we tried to limit the number of training steps for the
  hyperparameter search in 24 so that no algorithm would have an
  advantage of having more degrees of freedom to adjust to the data 
(but in section~\ref{sec:hyper} we discuss
  that some algorithms may allow testing many values of the
  hyperparameter with just one training and testing step). 

\item Besides computing when two algorithms are significantly
  different in their error rates, in the standard null-hypothesis
  significance tests (NHST), we also use a Bayesian analysis to
  discover when the differences of two algorithms has no ``practical
  consequences.''  As far as we know, this is the first time a
  Bayesian ANOVA is used to compare classifiers, which we
  believe is an important direction in analysis of results in
  empirical machine learning. But more significantly is the use of
  ``limits of practical significance'', that is, the definition of
  thresholds below which the differences are irrelevant from practical
  purposes, which goes beyond an more standard ``significance test''
  analysis currently used in the machine learning literature.

  \cite{delgado14} follow the standard null hypothesis significant
  test in analyzing their result, but even within this framework,
  their analysis is not as rigorous as it should have been. The NIST
  standard for comparing many classifiers across different datasets
  was proposed by \cite{demsar2006} and it is discussed in
  section~\ref{sec:demsar}. In particular, when testing the different
  algorithms for statistical significant differences, the Demsar
  procedure requires one to use the Nemenyi test, which is a
  nonparametric test that performs the appropriate multiple comparison
  p-value correction. But \cite{delgado14} used a paired t-test (a
  parametric test) between the first ranked and the following 9 top
  ranked algorithms, apparently without multiple comparions
  corrections. Very likely, given the large number of comparisons need
  to contrast all 179 algorithms, very few, if any of the pairwise
  comparisons would have been flagged as significant by the Nemenyi
  test.

In this paper we followed the full Demsar procedure to analyse the
results, but we also used the Bayesian ANOVA to verify when the
differences between the algoritms is not only ``statistically significant'',
but also ``of practical significance''.

\cite{delgado14} first two ranked programs are two different
implementation of the same algorithm - parallel implementation of
random forest  (first ranked) and a non-parallel implementation (second
ranked). The authors state that in some sense, the difference between
the two results should not be ``important''  (they use the term
``significant'') and indeed the statistical significance analysis
shows that the difference was not statistically significant, but
neither was the next 6 ranked algorithms (in comparison to the top
performing).

\item We studied the computational costs of running a ``standard''
  implementation of the different  algorithms. With information
  regarding the computational cost a practitioner may want to balance
  execution time and expected accuracy. Furthermore, this information may encourage
  the practitioner choose other implementations that the ones tested,
  and researcher to develop faster implementations of the best ranked
  algorithms.  

\end{itemize}

\section{Data and Methods}
\label{sec:data-methods}

\subsection{Experimental procedure}
\label{sec:exper-proc}

In general terms the experimental procedure followed by this research
is the following.

Each dataset $D_i$ (the datasets are discussed in
section~\ref{sec:datasets}) is divided in half into two subsets
$S_{1i}$ and $S_{2i}$, each with the same proportion of classes. For
each subset $S_{ji}$, we used a a 5-fold cross validation procedure to
select the best set of hyperparameters for the procedure $a$,
($\hat{\theta}_{a}$). Then we trained the whole subset $S_{ji}$ using
the procedure $a$ with hyperparameters $\hat{\theta}_j$ and computed
the error rate on the subset $S_{\jhat i}$, (where $\hat{1}=2$ and
$\hat{2}=1$). We call the error rate of algorithm $a$ when learning on
the subset $S_{ji}$, with hyperparameters $\hat{\theta}_{a}$, when
testing on the subset $S_{\jhat i}$ as
$\error(i,\jhat | a, \hat{\theta}, j)$.

The expected for procedure $a$ error on (whole) dataset $i$ is
$\error(i| a)$ and it is computed as
the average
\begin{equation}
  \label{eq:3}
  \error(i|a) = \frac{\error(i,1 | a, \hat{\theta}, i, 2)+ \error(i,2 | a, \hat{\theta}, i, 1)}{2}
\end{equation}
We then performed the analyses described in sections~\ref{sec:demsar} and
\ref{sec:banova} on the sets $\{ \error(i|a) \} $ for all datasets $i$ (described
in section~\ref{sec:datasets} and for all classification algorithms $a$ (section~\ref{sec:class-algor}).

\subsection{Datasets}
\label{sec:datasets}
We started with the 121 datasets collected from the UCI site,
processed, and converted by the authors of \cite{delgado14} into a
unified format. The datasets is derived from the 165 available at UCI
website in March 2013, where 56 were excluded by the authors of
\cite{delgado14}. For the remaining 121, \cite{delgado14} converted
all categorical variables to numerical data, and each feature was
standardized to zero mean and standard deviation equal to 1.

We downloaded the datasets preprocessed by the authors of
\cite{delgado14} in November 2014. We performed the following
transformations:
\begin{itemize}
\item 65 of the datasets were multiclass problems. We converted them
  to a binary classification problem by ordering the classes by their
  names, and alternatively assigning the original class to the
  positive and negative classes. The datasets have different
  proportions between the positive and negative classes, approximately
  normally distributed with mean 0.6 and standard deviation of 0.17.
\item 19 of the datasets were separated into different
  training and test sets and on the November 2014 data, the test set
  was not standardized. We standardized the test set (separately from
  the train set) and joined both sets to create a single dataset.
\item we removed the 6 datasets with less than 100 datapoints,
\item the 9 datasets with more than 10.000 datapoints (more data
  points for subset) we searched the hyperparameters on a subset of
  5000 datapoints. The final training was done with the whole subset,
  and the testing with the other full subset. 
\end{itemize}
Thus, in this research we tested the algorithms in 121 - 6 (very small
datasets removed)
= 115 datasets. 

Details of each dataset are described in \ref{ap:ds}.

\subsection{Classification algorithms}
\label{sec:class-algor}

We used 14 classification algorithms from very different families. As
discussed in the Introduction, we argued that one of the possible
criticisms to the \cite{delgado14} paper is that the authors do not
distinguish different algorithms from different implementations of
that algorithm. 

% Of course different implementation are particularly
% important in the execution time, and memory costs of running the
% algorithm. We also would include robustness to the important
% characteristic to distinguish different implementations - we did
% encounter implementations of algorithms that would generate exceptions
% and stop the execution for particular combinations of hyperparameters,
% and a few that would not run at all for a particular dataset.

% We believe that a first level of analysis that distinguished only
% between algorithms, and not necessarily between implementations, is
% more useful for both practitioners, and researchers. If one or a set of
% algorithms is shown to outperform the others, the practitioner should
% concentrate their efforts in these algorithms to solve their
% problems. For the researcher, they can

Although we do not have a clear or formal definition of what are
``families of algorithms'' we believe that we have a sufficiently 
diverse collection of algorithms.

We chose not to use algorithms that do not require hyperparameters,
such as Linear Discriminant Analysis (LDA) and logistic
regression. Hyperparameters allow the algorithm to better adjust to
the data details, and so we believe LDA and logistic regressions would
be in disadvantages to the other algorithm.  We added regularized
versions of these algorithms, which contain at least one
hyperparameters.

Thus, \texttt{glmnet} (L1 and L2 regularized logistic regression
\citep{zou2005regularization}) and \texttt{sda} a L1-regularized LDA
are two mainly linear algorithms. We would also add \texttt{bst} a
boosting of linear models.

Among the distance based classifiers, \texttt{knn}  is the usual
k-nearest neighbor, and \texttt{rknn} is a bagging of knn, where each
base learner is a \texttt{knn} on a random sample of the original
features. \texttt{lvq}, or learning vector quantization
\citep{kohonen1995learning} is a cluster plus distance, or dictionary
based classification: a set of ``prototypes,'' or clusters, or
``codebook'' is discovered in the data, many for each class, and  new
data is classified based on the distance to these prototypes. 

Neural network based classifiers include \texttt{nnet} a common
1-hidden layer logistic network, and \texttt{elm} or extreme learning
machines \citep{huang2006extreme}. Extreme learning machines are a
1-hidden layer neural network, where the weights from the input to the
hidden layer are set at random, and only the second set of weights are
learned (usually via least square optimization).

We only tested the naive Bayes \texttt{nb} as a Bayesian network based
algorithm.

We divided the SVM family into the linear SVM (\texttt{svmLinear}), the
polynomial SVM (\texttt{svmPoly}) and the RBF SVM
(\texttt{svmRadial}).

% We also included the Least-Square SVM
%\texttt{lssvmRadial}, a L2-regularized (instead of L1-regularized)
%SVM \cite{suykens1999least}.

Finally we included one implementation of random forests
(\textbf{rf}) and one implementation of gradient boosting machine
classifiers (\texttt{gbm}) \citep{friedman2001greedy}.

% We included two different variations of Random Forest. \texttt{rf} is
% the standard one, and \texttt{cforest} is a random forest of
% conditional inference trees \cite{hothorn2006unbiased,cfor2,cfor3}.
% Finally, we also included in the analysis two variations of a boosting
% of short decisions trees: \texttt{gbm} is gradient boosting machines classifier
% \cite{friedman2001greedy}  and \texttt{ada} a Adaboost (of short
% decisions trees) classifier. 
%The choice of the algorithms tested was informed by the results of
%\cite{delgado14}.
%Future work should probably include some 

The implementation information of the algorithms are listed below.

% Details on the
%hyperparameters of each algorithm is described in \ref{sec:class-algor}.

\begin{description}

% \item[ada] Boosting of decisions trees based on the Adaboost:
%   Implementation: package \emph{ada} \citep{adaR}).

\item[bst] Boosting of linear classifiers. We used the R
  implementation in the package \emph{bst} \citep{bstR}

%\item[cforest] Conditional random forest. Implementation: package \emph{party} %\citep{party}

\item[elm] Extreme learning machines Implementation:  package \emph{elmNN} \citep{elmNNR})

\item[gbm] Gradient boosting machines. Implementation: package \emph{gbm} \citep{gbmR}

\item[glmnet] Elastic net logistic regression classifier.
 Implementation : package \emph{glmnet} \citep{glmnetR})

\item[knn]  k-nearest neighbors classifier. Implementation: package \emph{class} \citep{classR}.

%\item[lssvmRadial] A least square SVM with Gaussian kernel  

\item[lvq] Learning vector quantization.
 Implementation: package \emph{class} \citep{classR})

\item[nb] Naive Bayes classifier: package \emph{klaR} \citep{klaRR}

\item[nnet] A 1-hidden layer neural network with sigmoid transfer
  function. Implementation: package \emph{nnet} \citep{classR}

\item[rf] Random forest. Implementation: package \emph{randomForest}
  \citep{randomForestR}

\item[rknn] A bagging of knn classifiers on a random subset of the
  original features.  Implementation: package \emph{rknn} \citep{rknn}

\item[sda] A L1 regularized linear discriminant
  classifier. Implementation: package \emph{sparseLDA}
  \citep{sparseLDAR}

\item[svmLinear] A SVM with linear kernel. package \emph{e1071} \citep{e1071R})

\item[svmPoly] A SVM with polynomial kernel. package \emph{e1071} \citep{e1071R})

\item[svmRadial] A SVM with RBF kernel. package \emph{e1071} \citep{e1071R})

\end{description}

\subsection{Hyperparameters ranges}
\label{sec:hyper}

The RandomForest classifier is a particularly convenient algorithm to
discuss the grid search on hyperparameters. Most implementation of
\texttt{rf} use two or three hyperparameters: the \texttt{mtry}, the
number of trees, and possible some limit of complexity of the trees,
either the minimum size of a terminal node, or the maximum profundity,
or a maximum number of terminal nodes. We did not set a range of
possible values for the hyperparameters that limit the complexity of
the trees. Mtry is the number of random features that will be used to
construct a particular tree. There is a general suggestion (we do not
know the source or other papers that tested this suggestion) that this
number should be the square root of the number of features of the
dataset. If nfeat is the number of features of the dataset, we set the
possible values to
$\{ 0.5 \times \sqrt{\mbox{nfeat}}, 1 \times \sqrt{\mbox{nfeat}}, 2
\times \sqrt{\mbox{nfeat}}, 3 \times \sqrt{\mbox{nfeat}}, 4 \times
\sqrt{\mbox{nfeat}}, 5 \times \sqrt{\mbox{nfeat}} \}$.
Also, the range is limited to at most nfeat$-1$.

The number of trees is what we will call a \textbf{free
  hyperparameter}, that is, an hyper-parameter can be tested for many
values but it only need one training step. One can train a random
forest with $n$ trees, but at testing time, some implementations can
return the classification of each tree on the test data. Thus one can
test the accuracy of a random forest with $m<n$ trees, just by
computing the more frequent classification of the first $m$ trees (or
any random sample of $m$ trees). Thus, to select this hyperparameter,
it is not necessary to create a grid with multiple training or
testing. So, for the random forest, \texttt{ntree} is a free hyperparameter,
that will be tested from 500 to 3000 by 500. But we also put another
limit on the number of trees, half of the number data points in the
subset.

The number of repetitions or boosts in a boosting procedure is also a
free parameter. The last class of free hyperparameter refer to an
implementation of classification algorithms that calculate all the
possible values of a parameter that causes changes in the learned
function. The only relevant algorithm for this research is the
elastic-net regularized logistic regression implemented by the package
glmnet \citep{glmnetR}), which computes all the values (or the
\emph{path} as it is called) of the regularization parameter
$\lambda$. \cite{hastie2004entire} discuss  a complete path algorithm
for SVM (for the C hyperparameter) but we did not use this
implementation in this paper. 

We list the range of hyperparameters for each of the algorithms
tested, where nfeat is the number of features in the data and ndat is
the number of data points.

\begin{description}

%\item[ada]  Implemented in R package \emph{ada}
% Hyperparameters, shrinkage = $\{ 0.05, 0.1 \}$, tree
%  depth = from 1 to 5, and then 4 values equally spaced from 7 to
% $nfeat-2$. Free hyperparameter, number of boosts: from 100 to 3000
% by 200, at most ndat.

\item[bst] Hyperparameters:  shrinkage =
  $\{ 0.05, 0.1\}$. Free hyperparameter, number of boosts, from 100 to
  3000 by 200, at most ndat.

%\item[cforest] Hyperparameters: mtry = $\{ 0.5, 1, 2, \}
 % \sqrt{nfeat}$ up to a value of $nfeat/2$. Number of trees is a free
 % hyperparameters, tested from 500 to 3000 by 500 up to $ndat/2$.

\item[elm]
Hyperparameter: number of hidden units = at most 24 values equally spaced
between $20$ and ndat$/2$

\item[gbm] Hyperparameters:
  interaction-depth = 1..5, shrinkage=$\{0.05,0.1\}$. number of boosts
  is a free hyperparameter, tested from 50 to 500 with increments of
  20 to at most ndat.

\item[glmnet] Hyperparameters
  $\alpha$, 8 values equally spaced between 0 and 1. Free
  hyperparameter $\lambda$, 20 values geometrically spaced between
  $10^{-5}$ to $10^3$.

\item[knn] Hyperparameter k:= 1,
  and at most 23 random values from 3 to ndat$/4$.

%\item[lssvmRadial] package \emph{kernlab} \citep{kernlab} Hyperparameters: C
 % = $2^-5, 2^0, 2^5, 2^{10}, 2^{15}$, $\gamma = 2^{-15}, 2^{-10.5}, 2^{-6},2^{-1.5}, 2^{3}$.

\item[lvq] Hyperparameter: size
  of the codebook = at most 8 values equally spaced between $2$ and
  $2 \times $ nfeat

\item[nb] Hyperparameters: usekernel = $\{$ true, false $\}$, fL = $\{
  0, 0.1, 1, 2 \}$

\item[nnet] Hyperparameter: number
  of hidden units = at most 8 values equally spaced between $3$ and
  nfeat$/2$, decay = $\{0, 0.01, 0.1\}$.

\item[rf] Hyperparameter mtry =
  $\{ 0.5,1,2,3,4,5 \} \sqrt{nfeat}$ up to a value of nfeat$/2$. Number of
  trees is a free hyperparameters, tested from 500 to 3000 by 500 up
  to ndat$/2$.

\item[rknn] Hyperparameters: mtry = 4 values
  equally spaced between 2 and nfeat$-2$, k= 1, and at most 23 random
  values from 3 to ndat$/4$. The number of classifiers is a free
  hyperparameter from 5 to 100, in steps of 20.

\item[sda] Hyperparameter: $\lambda$ = 8 values
  geometrically spaced from $10^{-8}$ to $10^{3}$
\item[svmLinear] . Hyperparameter: C
  = $2^{-5}, 2^0, 2^5, 2^{10}, 2^{15}$. 

\item[svmPoly] Hyperparameter C as in the linear
  kernel and degree from 2 to 5.

\item[svmRadial] Hyperparameters C as in the linear SVM
  and $\gamma = 2^{-15}, 2^{-10.5}, 2^{-6},2^{-1.5}, 2^{3}$.
\end{description}

\subsection{Reproducibility}
\label{sec:reproducibility}

The data described above, the R programs that tested the 14
algorithms, the results of running there algorithms, the R programs used
to analyse these results and generate the tables and figures in this
paper, and the saved interactions of the MCMC algorithm are available
at \url{https://figshare.com/s/d0b30e4ee58b3c9323fb}.

\section{Statistical procedures}
\label{compx}

% In this paper we performed three different statistical analysis. The
% first one will analise the difference $\delta_{ij} = \est(x,i,j)-\est(a,i,j)$, that
% is the accuracy gain in using the nested cv procedure to select the
% algorithm $x$, when the flat-cv procedure would have selected $a$. We
% will then show that aggregating all 

\subsection{Demsar procedure}\label{sec:demsar}

We will follow the procedure proposed by \cite{demsar2006}. The
procedure suggests one should first apply a Friedman test (which can
be seen as a non-parametric version of the repeated measure ANOVA
test) to determine if there is sufficient evidence that the error rate
measures for each procedure are not samples from the same
distribution. If the p-value is below 0.05 (for a 95\% confidence)
than one can claim that it is unlikely that the error rates are all
``the same'' and one can proceed to compare each algorithm to the
others. When comparing all algorithms among themselves, which is the
case here, Demsar proposed the Nemenyi test, which will compute the
p-value of all pairwise comparisons. Again, a p-value below 0.05
indicates that that comparison is statistically significant, that is,
it is unlikely that two sets fo error rates are samples from the same
distribution.

\subsection{Bayesian comparison of multiple groups}
\label{sec:banova}

A standard null hypothesis significant test assumes the null
hypothesis, usually that the two samples came form the same population
and computes the probability (p-value) of two samples from the same
population having as large a difference in mean (or median) as the one
encountered in the data. If the p-value is not low, one \textbf{cannot
  claim} that the null hypothesis is true and that the two data sets
came from the same population ( and thus all observed differences are
due to ``luck''). Failing to disprove the null hypothesis because the
p-value is too high is not the same as proving the null hypothesis.

Equivalence tests are a form of ``proving'' a weaker version of the
usual null hypothesis. Equivalence tests assume as the null hypothesis
that the difference between the mean (or median) of the two sets is
above a certain threshold, and if the p-value is low enough, one can
claim that this null hypothesis is false, and thus that the difference
between the means is smaller than the threshold. Equivalence tests are
useful when that threshold is a limit of \textbf{practical
  irrelevance}, that is, a limit below which changes in the mean of
two groups are of no practical consequence. Of course, this limit of
irrelevance is very problem dependent. Section~\ref{sec:irrelevance1}
will discuss our proposal for such a limit.

A different approach to prove that the differences are not important
is to use Bayesian methods. Bayesian methods will compute the
(posterior) distribution of probability for some set of measures,
given the prior hypothesis on those measures. Thus a Bayesian method
can compute the posterior distribution of the difference of two
means. The area of the distribution of the difference that falls
within the limits of irrelevance is the probability that the
difference is of no practical importance. In Bayesian statistics, the
limit of irrelevance is called Region of Practical Equivalence (ROPE).

% We will use  the Bayesian ANOVA procedure, as described in
% \cite{bayesbook}. In particular we will use a two factor ANOVA
% -- algorithm and dataset are the factors -- without interaction, where
% dataset is a subject factor (it indicates the subject or pairing in
% the ANOVA) and the algorithm is the factor we are interested in. 

The standard Bayesian ANOVA, as described in \cite{bayesbook} is based
on normal distributions. We are interested in a 2-factor ANOVA, where
one of the factors is the classification algorithm , and the other
factor is the dataset. We assume that there is no interaction
component, that is, we are not interested in determining
particularities of each algorithm on each dataset - we are making a
claim that the datasets used in this research are a sample of real
world problems, and we would like to make general statements about the
algorithms.

Let us denote $y_{ad}$ as the error rate for the algorithm $a$ on
dataset $d$, then the usual hierarchical model is \citep{bayesbook}:
\begin{subequations}
\begin{align}
  y_{ad} \sim & \mathbf{N}(\nu_{ad}, \sigma_0) \label{eq1}\\
\nu_{ad}  = & \beta + \alpha_a + \delta_d \label{eq8}\\
\sigma_0  \sim & \mathbf{U}(ySD/100,ySD*10 )\\
\beta  \sim & \mathbf{N}(yMean, ySD*5) \\
\alpha_a  \sim & \mathbf{N}(0, \sigma_a) \label{eq2}\\
\delta_d  \sim & \mathbf{N}(0,\sigma_d) \label{eq3}\\
\sigma_a  \sim & \text{Gamma}(ySD/2,ySD*2) \label{eqz4}\\
\sigma_d  \sim & \text{Gamma}(ySD/2,ySD*2) \label{eqz5}
\end{align}
\end{subequations}
where $\mathbf{U}(L,H)$ is the uniform distribution with $L$ and $H$
as the low and high limits; $\mathbf{N}(\mu,\sigma)$ is the normal
distribution with mean $\mu$ and standard deviation $\sigma$; and
Gamma($m, \sigma$) is the Gamma distribution with mode $m$ and
standard deviation $\sigma$ - note that this is not the usual parametrization of the
Gamma distribution. $ySD$ is the standard deviation of the $y_{ad}$
data, and $yMean$, the mean of that data.

We are interested in the joint posterior probability
$P(\alpha_1, \alpha_2, \ldots , \alpha_A | \{y_{ad}\}) $. From that
one can compute the relevant pairwise differences
$P(\alpha_i - \alpha_j| \{y_{ad}\}) $ and in particular, how much of
the probability mass fall within the region of irrelevant
differences. More specifically, if
$\vec{\alpha} = \langle \alpha_1, \alpha_2, \ldots , \alpha_A
\rangle$,
then the simulation approach we will follow generates a set
$\{ \vec{\alpha}_j | \vec{\alpha}_j \sim P(\vec{\alpha}| \{y_{ad}\})
\}$
for $j = 1 \ldots N$ where $N$ is the number of chains in the MCMC
simulation. We compute characteristics of distribution of
$P(\alpha_i - \alpha_j| \{y_{ad}\}) $ from the $\vec{\alpha}_j$.

Finally, one commonly used \emph{robust} variation of the model above
is to substitute the normal distribution in
Equation~\ref{eq1} for a student-t distribution, with low degree of
freedom, that is:
\begin{equation*}
 \begin{aligned}
 y_{ad} \sim & \mathbf{t}(\nu_{ad}, \sigma_0, df) \label{ex1}\\
df  \sim & \mbox{Exp}(1/30)
  \end{aligned}
\end{equation*}
where Exp$(\lambda)$ is the exponential distribution with rate
$\lambda$. We also run the robust version of the model, as discussed
in \ref{app:robust}.

\subsection{Threshold of irrelevance }
\label{sec:irrelevance1}

We propose two different forms of defining the threshold of
irrelevance for differences in error rates. We will compute the two
thresholds and use the lowest of the two measures as the threshold of
irrelevance.

The first proposal is to compare the two measures of error for each
dataset and for each classification algorithm. In Equation~\ref{eq:3}
they were the two terms $\error(i,1 | a, \hat{\theta}_1, i, 1)$ and
$\error(i,2 | a, \hat{\theta}_2, i, 2)$, that is, the error of
classifier $a$ when learning on training subset $S_{2i}$ and tested on
the subset $S_{1i}$, and the dual of that. The difference 
\begin{equation}
  \label{eq:9}
 \delta_{ai} = | \error(i,2 | a, \hat{\theta}_1, i, 1) - 
\error(i,1 | a, \hat{\theta}_2, i, 2) | 
\end{equation}
can be seen as the change on the error rate that one would expect from
applying the classifier $a$ on two different samples ($S_{1i}$ and
$S_{2i}$) from the same population ($D_{i}$). We will then compute
the median of $\delta_{ai} $ for all datasets $i$ and for all
classification algorithms $a$ that are among the three best (as
calculated by  $\error(i|a)$) for each dataset. That is, we are
considering as a threshold of irrelevance, the median of the change one
should expect from using a good classifier (among the top three for
that dataset) on two different samples of the same dataset.

The second proposal compares the estimate of the error
$\error(i,2 | a, \hat{\theta}, i, 1)$ computed from the 5-fold CV
procedure on the subset $S_{1i}$ which selects the best
hyperparameters with the measure of the error itself. We have not made
explicit the steps in the 5-fold CV, but intuitively, for each
combination of hyperparameters values $\theta$, it computes the
average of the error of training in 4 folds and testing on the
remaining one. Let us call it $\error_{cv}(i,1| a, \theta, i, 1)$ -
that is the CV error computed within the $S_{1i}$ subset itself.  We
select the combination of hyperparameters that have the lowest
$\error_{cv}(i,1| a, \theta, i, 1)$ . But this CV error is an estimate
of the future error of the classifier (trained with that combination
of hyperparameters). On the other hand
$\error(i,2 | a, \hat{\theta}, i, 1)$ is a ``future'' error of the
classifier - the training set is slightly different from the 5-fold CV
since it includes the whole subset $S_{1i}$ while for each fold it
included only 4/5 of $S_{1i}$, and the testing set $S_{2i}$ is totally
new. The difference
\begin{equation}
  \label{eq:10}
 \delta^{cv}_{ai} = | \error(i,2 | a, \hat{\theta}, i, 1) - \error_{cv}(i,1|  a,
\theta, i, 1)| 
\end{equation}
can be seen as the change in error one would expect from applying the
classifier $a$ on a slightly larger training set and testing it on
completely new data. We will compute the median of
$\delta^{cv}_{ai} $ for all datasets $i$ and for all classification
algorithms $a$ that are among the three best (as calculated by
$\error(i|a)$) for each dataset.

In both these proposals, we are defining the threshold of irrelevance
based on a ``futility'' point of view, and not based on theoretical
considerations. One knows that given a different sample from the same
distribution or given new data to test, all algorithms will have
different error rates.
 
\subsection{Computational costs}
\label{sec:computational-costs}

We compute to measures of computational costs to run the
algorithms. The first one is the \textbf{1-train-test}, which measures
the total time to train the algorithm $a$ on a subset $S_{ji}$ and to
test it in the subset $S_{\jhat i}$. Thus, that is the time to train
the algorithm and to run it on two equally sized data.

But all algorithms must search for the appropriate
hyperparameters. Thus also compute the total time to search for the
correct hyperparameter (using, as discusses above a 5-fold CV). But
different algorithms may have a different gird size of tentative
hyperparameters (since as discussed for some algorithms some of range
of hyperparameter may depend on characteristics of the dataset). Thus
we divide the total time to search for the best hyperparameter by the
number of hyperparameter combinations tested. We call it the
\textbf{per hyperparameter} time.

Since the execution time varies greatly on different datasets, we will
use the mean rank of each execution time to rank the algorithms, and
use the Demsar procedure to determine which execution times are
significantly different than the others (from a statistical sense). We
have no intuition on what could be considered an irrelevant change in
either execution times, so we will not perform the Bayesian ANOVA
analysis. 

The program ran on different cores of  a cluster of different
machines, so there is no guarantee that the machines have the same
hardware specification. But we ran all the 14 algorithms for a
particular subset on the same program. Thus the unit of distribution
is all the 14 algorithms searching for the best combination of
hyperparameters on a 5-fold CV of  half of a dataset, then training
on the full half dataset with the best selection of the hyperparameters
and finally testing the classifier on the other half of the
dataset.

Therefore, for the timing analysis we do not average the two measures
from the subsets to obtain the measure per dataset. Instead we perform
the Demsar statistical analysis using the subsets as subject
indicator.

\section{Results}
\label{sec:results}

\subsection{Error rates  of the different algorithms}
\label{sec:perf-diff-algor}

Table~\ref{tab1} list the mean ranking of all algorithms, the number
of times each algorithm was among the top for a dataset. The best
performing algorithm, in terms of mean rank across all subsets was the
random forest, followed by SVM with Gaussian kernels, followed by
gradient boosting machines. The three worst performing algorithms in
terms of mean rank were boosting of linear classifiers naive Bayes and
L1-regularized LDA. We make no claim that these three algorithms are
intrinsically ``bad''. It is possible that our choice of
hyperparameters was outside the range of more useful values, or the
implementation used was particularly not robust. In particular both
\texttt{nb} and \texttt{sda} did not run at all for 20 subsets, which
may explain partially their poor performance.

The ranking of the algorithm is dense, that is, all best performing
algorithm receive rank 1, all second best algorithms receive rank 2,
and so on. Also for the ranking, we rounded the error rates to 3
significant digits, so two algorithms have the same rank if their
error rates have a difference of less than 0.0005. We do not use the
rounding for the calculations of the irrelevance threshold.

% latex table generated in R 3.2.4 by xtable 1.8-2 package
% Sun May  8 13:18:25 2016
\begin{table}[ht]
\centering
\begin{tabular}{lrr}
  \hline
alg & mean rank & count \\ 
  \hline
  rf & 3.04 &  36 \\ 
  svmRadial & 3.36 &  24 \\ 
  gbm & 3.41 &  25 \\ 
  nnet & 4.88 &  13 \\ 
  rknn & 5.04 &  10 \\ 
  svmPoly & 5.14 &  11 \\ 
  knn & 5.32 &  11 \\ 
  svmLinear & 6.15 &  13 \\ 
  glmnet & 6.16 &  15 \\ 
  elm & 6.55 &   5 \\ 
  lvq & 6.96 &   8 \\ 
  sda & 7.05 &   5 \\ 
  nb & 8.23 &   7 \\ 
  bst & 9.08 &   4 \\ 
   \hline
\end{tabular}
\caption{The mean rank and the number of times the algorithm was among
  the top performer for each of the algorithms.}\label{tab1}
\end{table}

\begin{figure}[ht]
\begin{center}
\includegraphics[width=\textwidth]{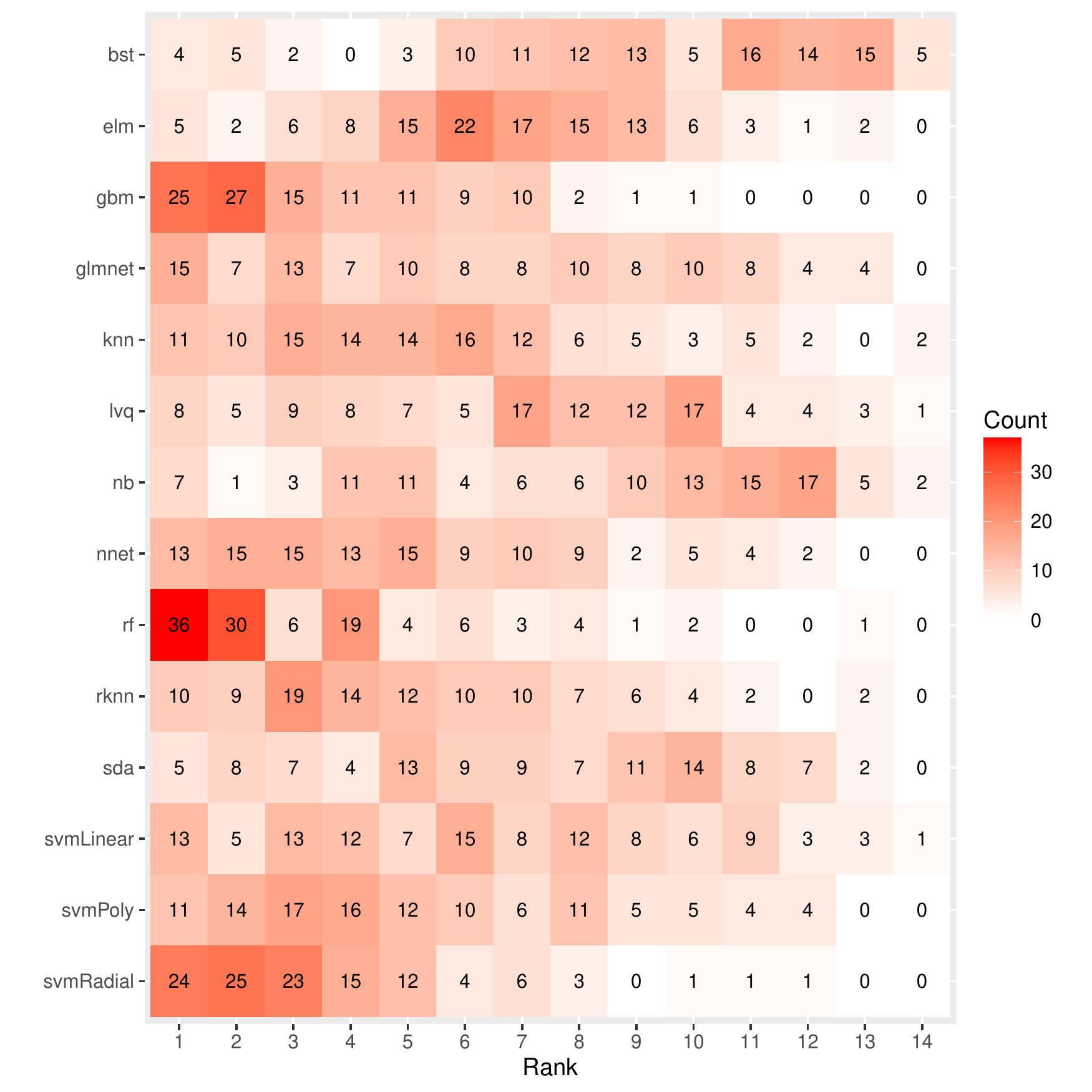}
\caption{The heat map with the distribution of the number of times each classifier achieved
  a particular rank. }\label{colormap}
\end{center}
\end{figure}

Figure~\ref{colormap}
displays the heat map of the rank distribution. 

\subsection{Demsar procedure}\label{demsarres}

Table~\ref{demsartab1} lists the p-value of the pairwise comparisons for all
classifiers. The p-values below 0.05  indicate that the
differences between the corresponding classifiers is statistically
significant.

% latex table generated in R 3.2.4 by xtable 1.8-2 package
% Sun May  8 13:20:06 2016
\begin{table}[ht]
\tiny
\centering
\begin{tabular}{rrrrrrrrrrrrrr}
  \hline
 & rf & svmRadial & gbm & nnet & rknn & svmPoly & knn & svmLinear & glmnet & elm & lvq & sda & nb \\ 
  \hline
svmRadial & 1.00 &  &  &  &  &  &  &  &  &  &  &  &  \\ 
  gbm & 1.00 & 1.00 &  &  &  &  &  &  &  &  &  &  &  \\ 
  nnet & \textbf{0.00} & \textbf{0.01} & \textbf{0.04} &  &  &  &  &  &  &  &  &  &  \\ 
  rknn & \textbf{0.00} & \textbf{0.00} & \textbf{0.00} & 1.00 &  &  &  &  &  &  &  &  &  \\ 
  svmPoly & \textbf{0.00} & \textbf{0.00} & \textbf{0.02} & 1.00 & 1.00 &  &  &  &  &  &  &  &  \\ 
  knn & \textbf{0.00} & \textbf{0.00} & \textbf{0.01} & 1.00 & 1.00 & 1.00 &  &  &  &  &  &  &  \\ 
  svmLinear & \textbf{0.00} & \textbf{0.00} & \textbf{0.00} & 0.75 & 1.00 & 0.86 & 0.93 &  &  &  &  &  &  \\ 
  glmnet & \textbf{0.00} & \textbf{0.00} & \textbf{0.00} & 0.73 & 1.00 & 0.84 & 0.92 & 1.00 &  &  &  &  &  \\ 
  elm & \textbf{0.00} & \textbf{0.00} & \textbf{0.00} & 0.07 & 0.54 & 0.12 & 0.19 & 1.00 & 1.00 &  &  &  &  \\ 
  lvq & \textbf{0.00} & \textbf{0.00} & \textbf{0.00} & \textbf{0.00} & 0.09 & \textbf{0.01} & \textbf{0.01} & 0.72 & 0.75 & 1.00 &  &  &  \\ 
  sda & \textbf{0.00} & \textbf{0.00} & \textbf{0.00} & \textbf{0.00} & \textbf{0.00} & \textbf{0.00} & \textbf{0.00} & 0.15 & 0.17 & 0.90 & 1.00 &  &  \\ 
  nb & \textbf{0.00} & \textbf{0.00} & \textbf{0.00} & \textbf{0.00} & \textbf{0.00} & \textbf{0.00} & \textbf{0.00} & \textbf{0.00} & \textbf{0.00} & 0.05 & 0.40 & 0.94 &  \\ 
  bst & \textbf{0.00} & \textbf{0.00} & \textbf{0.00} & \textbf{0.00} & \textbf{0.00} & \textbf{0.00} & \textbf{0.00} & \textbf{0.00} & \textbf{0.00} & \textbf{0.00} & \textbf{0.00} & \textbf{0.03} & 0.82 \\ 
   \hline
\end{tabular}
\caption{The p-values of the Nemenyi pairwise comparison
  procedure. P-values below 0.05 are in bold and denote that the
  difference in error rate of the two corresponding classifiers is
  statistically significant}\label{demsartab1}
\end{table}

The Demsar analysis shows that there is no statistical significance
difference between the top 3 classification algorithms -- the p-values
of the pairwise comparisons among \texttt{rf}, \texttt{svmRadial}, and
\texttt{gbm} are all well above 0.05. \texttt{nnet}, the next algorithm in
the rank, is statistically significantly different from each of the
first three. As discussed, the failure to prove that the three top
algorithms are not statistically different, does not prove that they
are equivalent. We will need the Bayesian ANOVA for that.

\subsection{Irrelevance thresholds}
\label{res.irrel}

The median $\delta_{ia}$ and $\delta^{cv}_{ia}$ for all datasets, and
for the three best algorithms for each dataset are: 
\begin{align*}
  \mbox{median}(\delta_{ia}) = &  0.0112\\
 \mbox{median}(\delta^{cv}_{ia}) = &  0.0134
\end{align*}
Thus, the two measures are comparable, a little over 1\%, and we will
use the lower among the two as out threshold of irrelevance
(0.0112). The fact that the median of  $\delta_{ia}$  is smaller than
the other is somewhat surprising, since  $\delta_{ia}$ measures the
error of learning (and testing) with two different samples of the data
population, while $\delta^{cv}_{ia}$ is the difference of learning on
basically the same subset, but testing in two different samples of
different sizes from the same data population.

\subsection{Bayesian ANOVA analysis}
\label{res.bayes}

% latex table generated in R 3.2.4 by xtable 1.8-2 package
% Tue May  3 11:18:33 2016
\begin{table}[ht]
\tiny
\centering
\begin{tabular}{rrrrrrrrrrrrrr}
  \hline
 & rf & svmRadial & gbm & nnet & rknn & svmPoly & knn & svmLinear & glmnet & elm & lvq & sda & nb \\ 
  \hline
svmRadial & 0.83 &  &  &  &  &  &  &  &  &  &  &  &  \\ 
  gbm & 0.74 & 0.64 &  &  &  &  &  &  &  &  &  &  &  \\ 
  nnet & 0.08 & 0.04 & 0.26 &  &  &  &  &  &  &  &  &  &  \\ 
  rknn & 0.01 & 0.00 & 0.04 & 0.62 &  &  &  &  &  &  &  &  &  \\ 
  svmPoly & 0.44 & 0.32 & 0.72 & 0.56 & 0.17 &  &  &  &  &  &  &  &  \\ 
  knn & 0.00 & 0.00 & 0.01 & 0.42 & 0.80 & 0.07 &  &  &  &  &  &  &  \\ 
  svmLinear & 0.00 & 0.00 & 0.00 & 0.20 & 0.61 & 0.02 & 0.78 &  &  &  &  &  &  \\ 
  glmnet & 0.00 & 0.00 & 0.00 & 0.02 & 0.20 & 0.00 & 0.36 & 0.61 &  &  &  &  &  \\ 
  elm & 0.00 & 0.00 & 0.00 & 0.06 & 0.33 & 0.00 & 0.52 & 0.74 & 0.82 &  &  &  &  \\ 
  lvq & 0.00 & 0.00 & 0.00 & 0.16 & 0.55 & 0.01 & 0.73 & 0.85 & 0.67 & 0.79 &  &  &  \\ 
  sda & 0.00 & 0.00 & 0.00 & 0.01 & 0.12 & 0.00 & 0.24 & 0.47 & 0.82 & 0.73 & 0.53 &  &  \\ 
  nb & 0.00 & 0.00 & 0.00 & 0.00 & 0.00 & 0.00 & 0.00 & 0.00 & 0.01 & 0.00 & 0.00 & 0.03 &  \\ 
  bst & 0.00 & 0.00 & 0.00 & 0.00 & 0.00 & 0.00 & 0.00 & 0.00 & 0.00 & 0.00 & 0.00 & 0.00 & 0.00 \\ 
   \hline
\end{tabular}
\caption{The probability that the difference between the error rate is
  within the limits of irrelevance (from -0.0112 to 0.0112) for all
  pairs of algorithms}\label{tabbanova}
\end{table}

Table~\ref{tabbanova} displays the pairwise probability that the
differences of the classifiers' error rates is within our limits of
irrelevance (from -0.0112 o 0.0112). 

The comparison between the top three algorithms shows that the
differences are with high probability (0.83 to 0.64)  within our range of irrelevance, that is, there
is no ``practical difference'' in the error rates of the three
algorithms.  In particular there is a stronger argument to be made
that the best algorithm, \texttt{rf} is equivalent for practical
purposed to the \texttt{svmRadial}. The claim of equivalence between
\texttt{gbm} and the other two is less strong, in particular, to our
surprise, in relation to the second best \texttt{svmRadial}.
% The result for the \texttt{nnet} shows an interesting
% phenomena. The Demsar analysis in Table~\ref{demsartab1} shows that
% the difference between \texttt{nnet} and the top 3 performing
% algorithms was statistically significant, but the data in
% Table~\ref{tabbanova} shows that there a 25\% probability the
%  difference between \texttt{nnet} and 
% \texttt{gbm} is of no practical consequence (93 and 95\% respectively)! That is, there is a
% ``real'' difference between the algorithms, in the sense that it is
% unlikely that the difference in results are only due to luck (or
% sampling error), but this real difference is probably of no practical
% consequence. 

\subsection{Computational costs of the algorithms}
\label{sec:comp-costs-algor}

% latex table generated in R 3.2.4 by xtable 1.8-2 package
% Sun May  8 13:27:04 2016
\begin{table}[ht]
\centering
\begin{tabular}{lrlr}
  \hline
alg & mean rank 1-t-t & alg & mean rank p/h \\ 
  \hline
  svmLinear & 2.67 & knn & 1.72 \\ 
  elm & 2.68 & lvq & 4.30 \\ 
  bst & 3.57 & \textbf{svmRadial} & 4.43 \\ 
  knn & 3.72 & sda & 4.56 \\ 
  sda & 4.21 & glmnet & 4.96 \\ 
  glmnet & 6.17 & svmPoly & 6.03 \\ 
  svmPoly & 7.16 & \textbf{gbm} & 6.65 \\ 
  lvq & 7.17 & nnet & 7.88 \\ 
  \textbf{svmRadial} & 7.73 & nb & 9.06 \\ 
  \textbf{gbm} & 8.88 & \textbf{rf} & 9.83 \\ 
  nnet & 9.54 & rknn & 10.70 \\ 
  \textbf{rf} & 10.94 & svmLinear & 11.25 \\ 
  rknn & 11.13 & elm & 11.67 \\ 
  nb & 12.19 & bst & 11.94 \\ 
   \hline
\end{tabular}
\caption{The mean rank for the 1-train-test measure (column 2: 1-t-t) and the
per hyperparameter measure (column 4: p/h)\label{tabtime}}
\end{table}

Table~\ref{tabtime} shows the mean rank for the 1-train-test and per
hyperparameter times. The top 6 algorithms in term of accuracy are
indeed computationally costly in terms of training and testing, and
they correspond to the 7 worse 1-train-test times. The result for the
\texttt{nb} is surprising. Also surprising, to the author at least, is
how more costly is \texttt{rf} in comparison to the two other
``equivalent'' classifiers, \texttt{svmRadial} and \texttt{gbm}.
Table~\ref{dem1trates} is the Demsar procedure p-values for the
pairwise comparisons of the 1-train-test cost of the top 6 algorithms
(\ref{fulltabs} contains the full p-value table). Notice that the
difference between \texttt{svmRadial} and \texttt{gbm} is not
significant; the difference between \texttt{rf} and \texttt{svmRadial}
and \texttt{gbm} are significant.

% The large per hyperparameter cost of \texttt{rf} may be
% explained by the fact that  number of trees is a free hyperparameter,
% and thus although there was only one training and testing steps (for
% which 
 Table~\ref{demperh} lists the p-values of the pairwise
comparisons of the top 6 algorithms.  \ref{fulltabs} contains the full p-value
table. \texttt{svmRadial} is significantly faster than all other algorithm in
the per hyperparameter measure.

\begin{table}[ht]
\centering
\begin{tabular}{rrrrrr}
  \hline
 & rf & svmRadial & gbm & nnet & rknn \\ 
  \hline
  svmRadial & \textbf{0.00} &  &  &  &  \\ 
  gbm & \textbf{0.01} & 0.69 &  &  &  \\ 
  nnet & 0.37 &\textbf{ 0.06} & 1.00 &  &  \\ 
  rknn & 1.00 & \textbf{0.00 }& \textbf{0.00} & 0.13 &  \\ 
  svmPoly &\textbf{ 0.00} & 1.00 & 0.08 & \textbf{0.00} & \textbf{0.00} \\ 
   \hline
\end{tabular}
\caption{The pairwise comparison p-value table for the top 6 algorithm of the 1-train-test computational cost of the algorithms.}\label{dem1trates}

\end{table}

% latex table generated in R 3.2.4 by xtable 1.8-2 package
% Sun May  8 13:32:25 2016
\begin{table}[ht]
\centering
\begin{tabular}{rrrrrr}
  \hline
 & rf & svmRadial & gbm & nnet & rknn \\ 
  \hline
svmRadial & \textbf{0.00} &  &  &  &  \\ 
  gbm & \textbf{0.00} &\textbf{ 0.00 }&  &  &  \\ 
  nnet & \textbf{0.03} & \textbf{0.00} & 0.61 &  &  \\ 
  rknn & 0.94 & \textbf{0.00} & \textbf{0.00} & \textbf{0.00} &  \\ 
  svmPoly & \textbf{0.00} & 0.17 & 1.00 & \textbf{0.05} & \textbf{0.00} \\ 
   \hline
\end{tabular}
\caption{The pairwise comparison p-value table for the top 6 algorithm  of the per hyperparameter cost of the algorithms.}\label{demperh}
\end{table}

The per hyperparameter cost is surprising low for the \texttt{svmRadial}, which
is a welcome result since the hyperparameter grid for the \texttt{svmRadial}
does not depend on characteristics of the dataset, and therefore one
will need to test many combinations, regardless of the
dataset. 

On the other hand the large per hyperparameter cost of
\texttt{rf} could be a problem if one cannot use the free
hyperparameter ``trick'' we described. Not all implementations of
\texttt{rf} allow access to each tree decision\footnote{The
  \texttt{predict} method of random forest in sklearn, for example,
  does not allow access to each three individual decision.}, and if one has to
explicitly train for different number of trees, the total number of
hyperparameters tested times the high per hyperparameter cost may make
the random forest solution less viable. 

\section{Discussion}
\label{sec:discussion}

An important part of these results is the definition of an irrelevance
threshold, a difference in error rates between to algorithms which is
likely to be irrelevant from practical purposes. We claim that 0.0112
is median change of error rate one should expect when a classifier is
trained and tested on a fully different set of data from the same
population.  This is a very strong claim, for which we have no
theoretical backing, but which is empirically derived from the 115
datasets used in this research. And if one agrees that these datasets
are a random sample of real life datasets that a practitioner will
``find in the wild'' (more on this issue below), that one must accept
that this conclusion is likely generalizable for all data.

In general terms, one should not loose too much sleep trying to improve a
classifier by less than 1\% since it is likely that one will loose (or
gain) that much when new data is used. 

We can provide some independent evidence that the change in error rate
with new data is around 1\%. Kaggle competitions are evaluated against
two hold out test sets, the public and the private. Usually the public
dataset is around 25 to 30\% of the private dataset (so the private
dataset is not totally new data, but 70 to 75\% new data. Thus the
difference of private and public accuracy scores are an independent
lower bound estimate of the change in accuracy for new data.
Unfortunately very few of the Kaggle competitions are measured using
accuracy, but one of them the ``cats and
dogs\footnote{\url{https://www.kaggle.com/c/dogs-vs-cats}}'' did. The
median of the difference between the public and private test scores,
for the top 50 entries on the public score is 0.0068, which is 60\% of
our threshold of irrelevance.

If one is reluctant to accept our 1\% as a limit of irrelevance when
comparing classifiers, Table~\ref{newbanova} shows the probability
that the difference between the accuracy rate is below 0.0056, that is
half of our irrelevance threshold, for the six best algorithms. Even
with this extremely rigorous limit of irrelevance, there is a some
probability that \texttt{rf}, \texttt{svmRadial}, and \texttt{gbm}
would perform at similar levels. In particular, there is still a 50\%
probability that  \texttt{rf} and \texttt{svmRadial} will be
equivalent for practical purposes. .

% latex table generated in R 3.2.4 by xtable 1.8-2 package
% Sun May  8 14:53:29 2016
\begin{table}[ht]
\centering
\begin{tabular}{rrrrrr}
  \hline
 & rf & svmRadial & gbm & nnet & rknn \\ 
  \hline
svmRadial & 0.51 &  &  &  &  \\ 
  gbm & 0.41 & 0.33 &  &  &  \\ 
  nnet & 0.02 & 0.01 & 0.08 &  &  \\ 
  rknn & 0.00 & 0.00 & 0.01 & 0.31 &  \\ 
  svmPoly & 0.18 & 0.11 & 0.40 & 0.26 & 0.05 \\ 
   \hline
\end{tabular}
\caption{Probability that the difference in error rates is within
  -0.0056 and 0.0056, for the six best ranking algorithms. }\label{newbanova}
\end{table}

Another important point of discussion is the Bayesian analysis itself. We
followed a more traditional modeling of the Bayesian ANOVA, where the
priors for the means follow a Gaussian distribution
(Equations~\ref{eq2} and \ref{eq3}). But error rates does not follow a
Gaussian distribution, first because they are limited to a 0-1
range. Second, the whole point of the Demsar procedure to compare
classifiers on different datasets is to use a non-parametric test,
one that  does not rely on the assumption of normality of the data. We
are not making an assumption that error rates are normally distributed,
but we are making an assumption that the priors for the coefficients
of the dependency on the dataset (given an algorithm) and on the
algorithm (given a dataset) are drawn from a Gaussian
distribution. \ref{bayescheck} discusses that those assumptions are
somewhat reasonable, but in the future, the community should probably
define a better model for the Bayesian analysis of error rates.
\ref{bayes2} shows that there was convergence for the Monte Carlo
simulations that generated the results in
Table~\ref{tabbanova}. Finally, we also ran the robust form of the
model. Table~\ref{newbanova} shows the probability that the
differences between the top six algorithms are within the limit of
practical relevance. Notice that one can be 100\% sure of the
equivalence of the top three algorithms using the robust model. Also
notice that even \texttt{nnet} which was significantly different from
the top three, is with high probability equivalent to at least
\texttt{svmRadial} and \texttt{gbm}. There is no contradiction in both
statements: the NHST claims that the difference is ``real'' while the
Bayesian ANOVA claims that that difference (although real) it is not
important. But nevertheless, the robust model seems to make much 
stronger claims of equivalence than the non-robust model, and as a
precaution, we preferred the latter model. The stronger results of the
robust model can be explained by shrinkage \citep{bayesbook}; since it allows
more outliers, the mean of each distribution of the algorithms'
coefficient would be ``free'' to move closer together, and thus the
higher probabilities that the differences are below the threshold of
irrelevance.

\begin{table}[ht]
\centering
\begin{tabular}{rrrrrr}
  \hline
 & rf & svmRadial & gbm & nnet & rknn \\ 
  \hline
svmRadial & 1.00 &  &  &  &  \\
  gbm & 1.00 & 1.00 &  &  &  \\ 
  nnet & 0.64 & 0.93 & 0.95 &  &  \\
  rknn & 0.74 & 0.96 & 0.97 & 1.00 &\\
  svmPoly & 0.50 & 0.85 & 0.88 & 1.00 & 1.00 \\
 \hline
\end{tabular}
\caption{Probability that the difference in error rates is within
  -0.0112 and 0.0112, for the six best ranking algorithms using the
  robust Bayesian model. }\label{newbanova}
\end{table}

\ref{app:robust} displays the full table for the pairwise
probabilities that the differences between the algorithms are within
the limit of irrelevance, and also discusses the convergency and model
checking of the robust model.

Our results are in general compatible with those in
\cite{delgado14}. Random forest is the best ranking algorithm in both
experiments; gradient boosting machines which was not included in
\cite{delgado14} performs well, as reported in various blogs, RBF SVM
also performs well. The divergence starts with the next best
classifications algorithms: \cite{delgado14} lists polynomial kernel
SVM \texttt{svmPoly} and extreme learning machines \texttt{elm} as the next best families, but in
our case \texttt{svmPoly} was equivalent to 1-hidden layer neural
nets \texttt{nnet}, and a bagging of knn \texttt{rknn}. The differences between these algorithms
was also below our practical limit of relevance. \texttt{elm} did
perform worse than \texttt{nnet} and \texttt{rknn}.

\subsection{Limits on this research}
\label{sec:limits-this-research}

One limit of this research is that its conclusions should only be
generalized to datasets ``similar'' to the ones used. In particular,
our datasets did not include very large, or very sparse, or datasets with
much more features than datapoints (as it is common in some
bioinformatics applications). Furthermore, our experiments were
performed on binary classification tasks. Given these restrictions on
the data, if one can assume that the datasets in UCI repository are
samples of ``real life'' problems, our results should be generalizable.

A second limit of this research is based on our decisions regarding
the hyperparameter search for each algorithm. There is very little
research on the range of possible or useful values of hyperparameters
for any of the algorithms discussed, so our decisions are
debatable. And if our choices were particularly bad, an algorithm may
have been unfairly ranked in the comparisons.

The reader should be aware of the limited usefulness for the timing
results. We used ``standard'' implementations of the algorithms in
R. All three implementations were written in some compiled language
and linked to R, but it may be that only libSvm, which is the base of
the SVM R implementation, has been under current development, and the
execution time of the SVM may be due to that. There has been more
recent implementations of random forest \citep{wright2015ranger} and
gradient boosting machines \citep{xgboost} which claim both a faster
execution time and shorter memory footprint. On the other hand,
incremental and online solvers for SVM
\citep{bordes2005fast,shalev2011pegasos} may further tip the scale in
favor of SVM.

\subsection{Future research}
\label{sec:future-research}

Given the result that \texttt{rf}, \texttt{svmRadial}, and
\texttt{gbm} are likely the best performing algorithms, it is
interesting to explore further variations and different
implementations of these algorithms. We discussed that more modern
implementations the algorithms may alter the timing results. But
variations of the algorithms may also alter the ranking and the
apparent equivalence of the three algorithms. Variations of Random
Forest, such as Rotation Forest \citep{rodriguez2006rotation},
Extremely Randomized Forest \citep{geurts2006extremely}, random forest of
conditional inference trees \citep{hothorn2006unbiased,cfor2}, should
be compared with the standard algorithm. Least square SVM \citep{suykens1999least}
should be compared with the standard SVM, and different models of
boosting such as AdaBoost, LogiBoost, BrownBoost, should be compared
with \texttt{gbm}.

There has been a large number of published research on different
methods and algorithms for hyperparameter selection in RBF SVM, but 
almost no research in hyperparameter selection for Random Forests and
Gradient Boosting Machines. Hyperparameter selection is a very
important and computationally expensive step in selecting an algorithm for a
particular problem, and further understanding of how to improve this
process is needed, specially for those two algorithms.

\section{Conclusion}
\label{sec:conclusion}

We have shown that random forests, RBF SVM, and gradient boosting
machines are classification algorithm that most likely will result in
the highest accuracy and that it is likely that there will be no
important difference in error rate among these three algorithms,
specially between random forest and RBF SVM. In terms of training and
testing execution times, SVM with a RBF kernel is faster then the two
other competitors.

We believe that this paper also makes important methodological
contributions for machine learning research, mainly in the definition
of a threshold of irrelevance, below which, changes in error rate
should be considered as of no practical significance. We argued that
this threshold should be 0.0112. Another important methodological
contribution is the use of a Bayesian Analysis of Variance method to
verify that the differences among the three top performing algorithms,
was very likely smaller than the 0.0112 threshold of practical
irrelevance, and we showed that the Bayesian model used is appropriate
to be used in comparisons of error rates among different algorithms.

%\bibliographystyle{elsarticle-harv}
%\bibliography{bib/r,bib/extra,bib/sites,bib/cv,bib/svm1,bib/classalg.bib}

\begin{thebibliography}{35}
\expandafter\ifx\csname natexlab\endcsname\relax\def\natexlab#1{#1}\fi
\expandafter\ifx\csname url\endcsname\relax
  \def\url#1{\texttt{#1}}\fi
\expandafter\ifx\csname urlprefix\endcsname\relax\def\urlprefix{URL }\fi

\bibitem[{Bordes et~al.(2005)Bordes, Ertekin, Weston, and
  Bottou}]{bordes2005fast}
Bordes, A., Ertekin, S., Weston, J., Bottou, L., 2005. Fast kernel classifiers
  with online and active learning. The Journal of Machine Learning Research 6,
  1579--1619.

\bibitem[{Brooks and Gelman(1998)}]{brooks1998general}
Brooks, S., Gelman, A., 1998. General methods for monitoring convergence of
  iterative simulations. Journal of Computational and Graphical Statistics
  7~(4), 434--455.

\bibitem[{Clemmensen(2012)}]{sparseLDAR}
Clemmensen, L., 2012. sparseLDA: Sparse Discriminant Analysis. R package
  version 0.1-6.
\newline\urlprefix\url{http://CRAN.R-project.org/package=sparseLDA}

\bibitem[{Demsar(2006)}]{demsar2006}
Demsar, J., 2006. Statistical comparisons of classifiers over multiple data
  sets. The Journal of Machine Learning Research 7, 1--30.

\bibitem[{Dietterich and Bakiri(1995)}]{dietterich1995solving}
Dietterich, T., Bakiri, G., 1995. Solving multiclass learning problems via
  error-correcting output codes. Journal of Artificial Intelligence Research,
  263--286.

\bibitem[{{Distributed (Deep) Machine Learning Community}(2016)}]{xgboost}
{Distributed (Deep) Machine Learning Community}, 2016. {xgboost: Scalable,
  Portable and Distributed Gradient Boosting (GBDT, GBRT or GBM) Library, for
  Python, R, Java, Scala, C++ and more}. \url{https://github.com/dmlc/xgboost}.

\bibitem[{Engel(1988)}]{engel1988polytomous}
Engel, J., 1988. Polytomous logistic regression. Statistica Neerlandica 42~(4),
  233--252.

\bibitem[{Fern\'{a}ndez-Delgado et~al.(2014)Fern\'{a}ndez-Delgado, Cernadas,
  Barro, and Amorim}]{delgado14}
Fern\'{a}ndez-Delgado, M., Cernadas, E., Barro, S., Amorim, D., 2014. Do we
  need hundreds of classifiers to solve real world classification problems?
  Journal of Machine Learning Research 15, 3133--3181.

\bibitem[{Franc et~al.(2002)}]{franc2002multi}
Franc, V., et~al., 2002. Multi-class support vector machine. In: 16th
  International Conference on Pattern Recognition. Vol.~2. pp. 236--239.

\bibitem[{Friedman et~al.(2010)Friedman, Hastie, and Tibshirani}]{glmnetR}
Friedman, J., Hastie, T., Tibshirani, R., 2010. Regularization paths for
  generalized linear models via coordinate descent. Journal of statistical
  software 33~(1), 1.

\bibitem[{Friedman(2001)}]{friedman2001greedy}
Friedman, J.~H., 2001. Greedy function approximation: a gradient boosting
  machine. Annals of Statistics, 1189--1232.

\bibitem[{Gelman et~al.(1996)Gelman, Meng, and Stern}]{gelman1996posterior}
Gelman, A., Meng, X.-L., Stern, H., 1996. Posterior predictive assessment of
  model fitness via realized discrepancies. Statistica Sinica, 733--760.

\bibitem[{Geurts et~al.(2006)Geurts, Ernst, and Wehenkel}]{geurts2006extremely}
Geurts, P., Ernst, D., Wehenkel, L., 2006. Extremely randomized trees. Machine
  Learning 63~(1), 3--42.

\bibitem[{Gosso(2012)}]{elmNNR}
Gosso, A., 2012. elmNN: Implementation of ELM (Extreme Learning Machine)
  algorithm for SLFN (Single Hidden Layer Feedforward Neural Networks). R
  package version 1.0.
\newline\urlprefix\url{http://CRAN.R-project.org/package=elmNN}

\bibitem[{Hastie et~al.(2004)Hastie, Rosset, Tibshirani, and
  Zhu}]{hastie2004entire}
Hastie, T., Rosset, S., Tibshirani, R., Zhu, J., 2004. The entire
  regularization path for the support vector machine. The Journal of Machine
  Learning Research 5, 1391--1415.

\bibitem[{Hastie et~al.(1998)Hastie, Tibshirani,
  et~al.}]{hastie1998classification}
Hastie, T., Tibshirani, R., et~al., 1998. Classification by pairwise coupling.
  The Annals of Statistics 26~(2), 451--471.

\bibitem[{Hothorn et~al.(2006)Hothorn, Hornik, and
  Zeileis}]{hothorn2006unbiased}
Hothorn, T., Hornik, K., Zeileis, A., 2006. Unbiased recursive partitioning: A
  conditional inference framework. Journal of Computational and Graphical
  Statistics 15~(3), 651--674.

\bibitem[{Huang et~al.(2006)Huang, Zhu, and Siew}]{huang2006extreme}
Huang, G.-B., Zhu, Q.-Y., Siew, C.-K., 2006. Extreme learning machine: theory
  and applications. Neurocomputing 70~(1), 489--501.

\bibitem[{Kohonen(1995)}]{kohonen1995learning}
Kohonen, T., 1995. Learning Vector Quantization. Springer.

\bibitem[{Kruschke(2014)}]{bayesbook}
Kruschke, J., 2014. Doing Bayesian data analysis: A tutorial with R, JAGS, and
  Stan. Academic Press.

\bibitem[{Li(2015)}]{rknn}
Li, S., 2015. rknn: Random KNN Classification and Regression. R package version
  1.2-1.
\newline\urlprefix\url{https://cran.r-project.org/web/packages/rknn/index.html}

\bibitem[{Liaw and Wiener(2002)}]{randomForestR}
Liaw, A., Wiener, M., 2002. Classification and regression by randomforest. R
  News 2~(3), 18--22.

\bibitem[{Meyer et~al.(2014)Meyer, Dimitriadou, Hornik, Weingessel, and
  Leisch}]{e1071R}
Meyer, D., Dimitriadou, E., Hornik, K., Weingessel, A., Leisch, F., 2014.
  e1071: Misc Functions of the Department of Statistics (e1071), TU Wien. R
  package version 1.6-4.
\newline\urlprefix\url{http://CRAN.R-project.org/package=e1071}

\bibitem[{Ridgeway(2013)}]{gbmR}
Ridgeway, G., 2013. gbm: Generalized Boosted Regression Models. R package
  version 2.1.
\newline\urlprefix\url{http://CRAN.R-project.org/package=gbm}

\bibitem[{Rodriguez et~al.(2006)Rodriguez, Kuncheva, and
  Alonso}]{rodriguez2006rotation}
Rodriguez, J.~J., Kuncheva, L.~I., Alonso, C.~J., 2006. Rotation forest: A new
  classifier ensemble method. IEEE Transactions on Pattern Analysis and Machine
  Intelligence 28~(10), 1619--1630.

\bibitem[{Shalev-Shwartz et~al.(2011)Shalev-Shwartz, Singer, Srebro, and
  Cotter}]{shalev2011pegasos}
Shalev-Shwartz, S., Singer, Y., Srebro, N., Cotter, A., 2011. Pegasos: Primal
  estimated sub-gradient solver for {SVM}. Mathematical Programming 127~(1),
  3--30.

\bibitem[{Strobl et~al.(2007)Strobl, Boulesteix, Zeileis, and Hothorn}]{cfor2}
Strobl, C., Boulesteix, A.-L., Zeileis, A., Hothorn, T., 2007. Bias in random
  forest variable importance measures: Illustrations, sources and a solution.
  BMC Bioinformatics 8~(25).

\bibitem[{Suykens and Vandewalle(1999)}]{suykens1999least}
Suykens, J.~A., Vandewalle, J., 1999. Least squares support vector machine
  classifiers. Neural Processing Letters 9~(3), 293--300.

\bibitem[{Venables and Ripley(2002)}]{classR}
Venables, W.~N., Ripley, B.~D., 2002. Modern Applied Statistics with S, 4th
  Edition. Springer, New York, iSBN 0-387-95457-0.
\newline\urlprefix\url{http://www.stats.ox.ac.uk/pub/MASS4}

\bibitem[{Wang(2014)}]{bstR}
Wang, Z., 2014. bst: Gradient Boosting. R package version 0.3-4.
\newline\urlprefix\url{http://CRAN.R-project.org/package=bst}

\bibitem[{Weihs et~al.(2005)Weihs, Ligges, Luebke, and Raabe}]{klaRR}
Weihs, C., Ligges, U., Luebke, K., Raabe, N., 2005. klar analyzing german
  business cycles. In: Baier, D., Decker, R., Schmidt-Thieme, L. (Eds.), Data
  Analysis and Decision Support. Springer-Verlag, Berlin, pp. 335--343.

\bibitem[{Wolpert(1992)}]{wolpert1992stacked}
Wolpert, D.~H., 1992. Stacked generalization. Neural Networks 5~(2), 241--259.

\bibitem[{Wright and Ziegler(2015)}]{wright2015ranger}
Wright, M.~N., Ziegler, A., 2015. ranger: A fast implementation of random
  forests for high dimensional data in {C++ and R}. arXiv:1508.04409.

\bibitem[{Zajac(2016)}]{delgado-gbm}
Zajac, Z., 2016. What is better: gradient-boosted trees, or a random forest?
  \url{http://fastml.com/what-is-better-gradient-boosted-trees-or-random-forest/}.

\bibitem[{Zou and Hastie(2005)}]{zou2005regularization}
Zou, H., Hastie, T., 2005. Regularization and variable selection via the
  elastic net. Journal of the Royal Statistical Society: Series B 67~(2),
  301--320.

\end{thebibliography}

 \appendix
\section{Datasets}
\label{ap:ds}

The table below list the characteristics of all datasets, ordered by size. The name of
the dataset is the same as the ones used in \cite{delgado14}. The size
refers to one half of the dataset. The
\emph{nfeat} column is the number of features of the datasets;
\emph{ndat} the number of data; \emph{prop} the proportion of
data of the positive class. \emph{Notes} has the following values:
\begin{itemize}
\item \emph{m} the dataset was multivalued, and so was converted to a
  binary problem using the procedure discussed in section~\ref{sec:data-methods}
\item \emph{l} large dataset. Only 5000 of the data was used
  to search fror the hyperparameters (see section~\ref{sec:data-methods}).
\end{itemize}

%Figure~\ref{fighist} displays the histogram of the proportion of the
%positive class. 

% latex table generated in R 3.1.2 by xtable 1.7-4 package
% Wed Dec 31 05:40:30 2014
%\begin{table}[ht]
\begin{center}
\begin{longtable}{lrrrr}
oocytes\_merluccius\_states\_2f &  26 & 511 & 0.93 & m \kill
\caption{The datasets}\\
\hline
dataset & nfeat & ndat & prop & note\\ 
\hline
\endhead
\hline
continued in the next page
\endfoot
\hline
\endlastfoot
  fertility &  10 &  50 & 0.96 & \\ 
  zoo &  17 &  50 & 0.62 & m\\ 
  pittsburg-bridges-REL-L &   8 &  51 & 0.82 &m \\ 
  pittsburg-bridges-T-OR-D &   8 &  51 & 0.86 &\\ 
  pittsburg-bridges-TYPE &   8 &  52 & 0.62 & m\\ 
  breast-tissue &  10 &  53 & 0.53 &m \\ 
  molec-biol-promoter &  58 &  53 & 0.49 & \\ 
  pittsburg-bridges-MATERIAL &   8 &  53 & 0.94& m \\ 
  acute-inflammation &   7 &  60 & 0.52 & \\ 
  acute-nephritis &   7 &  60 & 0.67 & \\ 
  heart-switzerland &  13 &  61 & 0.38 & m \\ 
  echocardiogram &  11 &  65 & 0.74 & \\ 
  lymphography &  19 &  74 & 0.46 & m\\ 
  iris &   5 &  75 & 0.72 & m\\ 
  teaching &   6 &  75 & 0.64& m \\ 
  hepatitis &  20 &  77 & 0.22 &\\ 
  hayes-roth &   4 &  80 & 0.57 & m \\ 
  wine &  14 &  89 & 0.60 & m\\ 
  planning &  13 &  91 & 0.74 & \\ 
  flags &  29 &  97 & 0.47 &m \\ 
  parkinsons &  23 &  97 & 0.25& \\ 
  audiology-std &  60 &  98 & 0.64 & m \\ 
  breast-cancer-wisc-prog &  34 &  99 & 0.77 & \\ 
  heart-va &  13 & 100 & 0.52 & m\\ 
  conn-bench-sonar-mines-rocks &  61 & 104 & 0.54 & \\ 
  seeds &   8 & 105 & 0.64 & m \\ 
  glass &  10 & 107 & 0.40 & m\\ 
  spect &  23 & 132 & 0.67 & m \\ 
  spectf &  45 & 133 & 0.19 & \\ 
  statlog-heart &  14 & 135 & 0.53 &  \\ 
  breast-cancer &  10 & 143 & 0.69 & \\ 
  heart-hungarian &  13 & 147 & 0.63&  \\ 
  heart-cleveland &  14 & 151 & 0.75 & m \\ 
  haberman-survival &   4 & 153 & 0.75 & \\ 
  vertebral-column-2clases &   7 & 155 & 0.68 & \\ 
  vertebral-column-3clases &   7 & 155 & 0.67 & m \\ 
  primary-tumor &  18 & 165 & 0.68 & m\\ 
  ecoli &   8 & 168 & 0.64 & m \\ 
  ionosphere &  34 & 175 & 0.30 &  \\ 
  libras &  91 & 180 & 0.51 & m \\ 
  dermatology &  35 & 183 & 0.64 & m  \\ 
  horse-colic &  26 & 184 & 0.64 & \\ 
  congressional-voting &  17 & 217 & 0.59 & \\ 
  arrhythmia & 263 & 226 & 0.67 & m \\ 
  musk-1 & 167 & 238 & 0.58 & \\ 
  cylinder-bands &  36 & 256 & 0.38&  \\ 
  low-res-spect & 101 & 265 & 0.25 & m \\ 
  monks-3 &   7 & 277 & 0.48 & \\ 
  monks-1 &   7 & 278 & 0.51 & \\ 
  breast-cancer-wisc-diag &  31 & 284 & 0.63 & \\ 
  ilpd-indian-liver &  10 & 291 & 0.72 & \\ 
  monks-2 &   7 & 300 & 0.64 & \\ 
  synthetic-control &  61 & 300 & 0.51 & m  \\ 
  balance-scale &   5 & 312 & 0.53 & m \\ 
  soybean &  36 & 341 & 0.42 & m \\ 
  credit-approval &  16 & 345 & 0.43 &  \\ 
  statlog-australian-credit &  15 & 345 & 0.32 & \\ 
  breast-cancer-wisc &  10 & 349 & 0.66 & \\ 
  blood &   5 & 374 & 0.76 & \\ 
  energy-y1 &   9 & 384 & 0.80 & m  \\ 
  energy-y2 &   9 & 384 & 0.75 & m \\ 
  pima &   9 & 384 & 0.66 & \\ 
  statlog-vehicle &  19 & 423 & 0.52 & m  \\ 
  annealing &  32 & 449 & 0.81 & m \\ 
  oocytes\_trisopterus\_nucleus\_2f &  26 & 456 & 0.41 &  \\ 
  oocytes\_trisopterus\_states\_5b &  33 & 456 & 0.98 & m\\ 
  tic-tac-toe &  10 & 479 & 0.34 & \\ 
  mammographic &   6 & 480 & 0.56 &  \\ 
  conn-bench-vowel-deterding &  12 & 495 & 0.53 & m \\ 
  led-display &   8 & 500 & 0.51 & m \\ 
  statlog-german-credit &  25 & 500 & 0.72 &  \\ 
  oocytes\_merluccius\_nucleus\_4d &  42 & 511 & 0.31 & \\ 
  oocytes\_merluccius\_states\_2f &  26 & 511 & 0.93 & m \\ 
  hill-valley & 101 & 606 & 0.49 & \\ 
  contrac &  10 & 736 & 0.79 & m \\ 
  yeast &   9 & 742 & 0.55 & m \\ 
  semeion & 257 & 796 & 0.50 & m \\ 
  plant-texture &  65 & 799 & 0.50 & m \\ 
  wine-quality-red &  12 & 799 & 0.56 & m \\ 
  plant-margin &  65 & 800 & 0.52 & m \\ 
  plant-shape &  65 & 800 & 0.52 & m \\ 
  car &   7 & 864 & 0.28 & m \\ 
  steel-plates &  28 & 970 & 0.66 & m  \\ 
  cardiotocography-10clases &  22 & 1063 & 0.39 & m  \\ 
  cardiotocography-3clases &  22 & 1063 & 0.86 & m \\ 
  titanic &   4 & 1100 & 0.66 & \\ 
  image-segmentation &  19 & 1155 & 0.58 & m \\ 
  statlog-image &  19 & 1155 & 0.55 & m  \\ 
  ozone &  73 & 1268 & 0.97 & \\ 
  molec-biol-splice &  61 & 1595 & 0.75 & m \\ 
  chess-krvkp &  37 & 1598 & 0.48 & \\ 
  abalone &   9 & 2088 & 0.68 & m\\ 
  bank &  17 & 2260 & 0.88 & \\ 
  spambase &  58 & 2300 & 0.61 & \\ 
  wine-quality-white &  12 & 2449 & 0.48 & m\\ 
  waveform-noise &  41 & 2500 & 0.66 & m\\ 
  waveform &  22 & 2500 & 0.68 & m\\ 
  wall-following &  25 & 2728 & 0.78 &m  \\ 
  page-blocks &  11 & 2736 & 0.92 & m \\ 
  optical &  63 & 2810 & 0.51 & m \\ 
  statlog-landsat &  37 & 3217 & 0.56 \\ 
  musk-2 & 167 & 3299 & 0.85 & m \\ 
  thyroid &  22 & 3600 & 0.94 & m \\ 
  ringnorm &  21 & 3700 & 0.49 & \\ 
  twonorm &  21 & 3700 & 0.49 &\\ 
  mushroom &  22 & 4062 & 0.51 &  \\ 
  pendigits &  17 & 5496 & 0.51 &ml\\ 
  nursery &   9 & 6480 & 0.68 &ml\\ 
  magic &  11 & 9510 & 0.65 & l\\ 
  letter &  17 & 10000 & 0.50 & ml\\ 
  chess-krvk &   7 & 14028 & 0.53 &ml \\ 
  adult &  15 & 24421 & 0.76 &l\\ 
  statlog-shuttle &  10 & 29000 & 0.84 &ml \\ 
  connect-4 &  43 & 33778 & 0.75 & l\\ 
 miniboone &  51 & 65032 & 0.28 & l\\ 
\end{longtable}
\end{center}

% \begin{figure}[t]
% \begin{center}
% \includegraphics[width=0.7\textwidth]{histpos.pdf}
% \caption{Histogram of the proportion of the positive class for the 115
%   datasets analyzed.\label{fighist}}
% \end{center}
% \end{figure}

 \section{Full p-value tables for the pairwise comparison of the
   1-train-test and per hyperparameter costs}
\label{fulltabs}

\begin{table}[ht]
\centering 
\tiny
\begin{tabular}{rrrrrrrrrrrrrr}
  \hline
 & rf & svmRadial & gbm & nnet & rknn & svmPoly & knn & svmLinear & glmnet & elm & lvq & sda & nb \\ 
  \hline
svmRadial & 0.00 &  &  &  &  &  &  &  &  &  &  &  &  \\ 
  gbm & 0.01 & 0.69 &  &  &  &  &  &  &  &  &  &  &  \\ 
  nnet & 0.37 & 0.06 & 1.00 &  &  &  &  &  &  &  &  &  &  \\ 
  rknn & 1.00 & 0.00 & 0.00 & 0.13 &  &  &  &  &  &  &  &  &  \\ 
  svmPoly & 0.00 & 1.00 & 0.08 & 0.00 & 0.00 &  &  &  &  &  &  &  &  \\ 
  knn & 0.00 & 0.00 & 0.00 & 0.00 & 0.00 & 0.00 &  &  &  &  &  &  &  \\ 
  svmLinear & 0.00 & 0.00 & 0.00 & 0.00 & 0.00 & 0.00 & 0.63 &  &  &  &  &  &  \\ 
  glmnet & 0.00 & 0.20 & 0.00 & 0.00 & 0.00 & 0.90 & 0.00 & 0.00 &  &  &  &  &  \\ 
  elm & 0.00 & 0.00 & 0.00 & 0.00 & 0.00 & 0.00 & 0.66 & 1.00 & 0.00 &  &  &  &  \\ 
  lvq & 0.00 & 1.00 & 0.08 & 0.00 & 0.00 & 1.00 & 0.00 & 0.00 & 0.89 & 0.00 &  &  &  \\ 
  sda & 0.00 & 0.00 & 0.00 & 0.00 & 0.00 & 0.00 & 1.00 & 0.07 & 0.01 & 0.08 & 0.00 &  &  \\ 
  nb & 0.60 & 0.00 & 0.00 & 0.00 & 0.89 & 0.00 & 0.00 & 0.00 & 0.00 & 0.00 & 0.00 & 0.00 &  \\ 
  bst & 0.00 & 0.00 & 0.00 & 0.00 & 0.00 & 0.00 & 1.00 & 0.91 & 0.00 & 0.92 & 0.00 & 0.96 & 0.00 \\ 
   \hline
\end{tabular}
\caption{The full p-value table for the 1-train-test time}
\end{table}

% latex table generated in R 3.2.4 by xtable 1.8-2 package
% Sun May  8 13:31:17 2016
\begin{table}[ht]
\centering
\tiny
\begin{tabular}{rrrrrrrrrrrrrr}
  \hline
 & rf & svmRadial & gbm & nnet & rknn & svmPoly & knn & svmLinear & glmnet & elm & lvq & sda & nb \\ 
  \hline
svmRadial & 0.00 &  &  &  &  &  &  &  &  &  &  &  &  \\ 
  gbm & 0.00 & 0.00 &  &  &  &  &  &  &  &  &  &  &  \\ 
  nnet & 0.03 & 0.00 & 0.61 &  &  &  &  &  &  &  &  &  &  \\ 
  rknn & 0.94 & 0.00 & 0.00 & 0.00 &  &  &  &  &  &  &  &  &  \\ 
  svmPoly & 0.00 & 0.17 & 1.00 & 0.05 & 0.00 &  &  &  &  &  &  &  &  \\ 
  knn & 0.00 & 0.00 & 0.00 & 0.00 & 0.00 & 0.00 &  &  &  &  &  &  &  \\ 
  svmLinear & 0.35 & 0.00 & 0.00 & 0.00 & 1.00 & 0.00 & 0.00 &  &  &  &  &  &  \\ 
  glmnet & 0.00 & 1.00 & 0.11 & 0.00 & 0.00 & 0.80 & 0.00 & 0.00 &  &  &  &  &  \\ 
  elm & 0.05 & 0.00 & 0.00 & 0.00 & 0.92 & 0.00 & 0.00 & 1.00 & 0.00 &  &  &  &  \\ 
  lvq & 0.00 & 1.00 & 0.00 & 0.00 & 0.00 & 0.09 & 0.00 & 0.00 & 1.00 & 0.00 &  &  &  \\ 
  sda & 0.00 & 1.00 & 0.01 & 0.00 & 0.00 & 0.29 & 0.00 & 0.00 & 1.00 & 0.00 & 1.00 &  &  \\ 
  nb & 0.98 & 0.00 & 0.00 & 0.67 & 0.13 & 0.00 & 0.00 & 0.01 & 0.00 & 0.00 & 0.00 & 0.00 &  \\ 
  bst & 0.01 & 0.00 & 0.00 & 0.00 & 0.64 & 0.00 & 0.00 & 0.99 & 0.00 & 1.00 & 0.00 & 0.00 & 0.00 \\ 
   \hline
\end{tabular}
 \caption{The full p-value table for the per hyperparameter times}
\end{table}

\section{Bayesian model verification}
\label{bayescheck}

\begin{figure}[t]
\begin{center}
\includegraphics[width=0.7\textwidth]{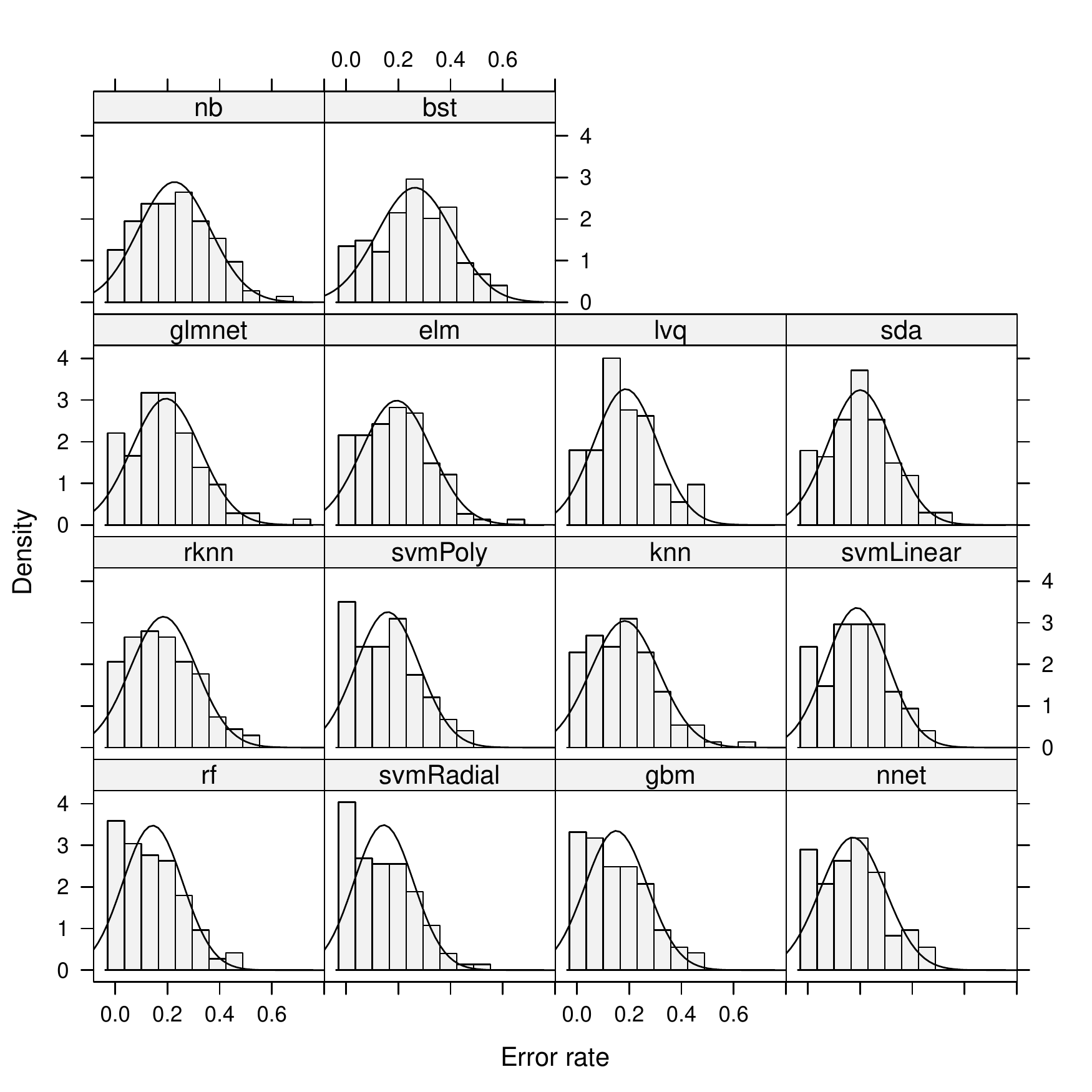}
\caption{Histogram of error rates for each algorithm across the
  different datasets, and the best fit Gaussian distribution.
\label{hist1}}
\end{center}
\end{figure}

\begin{figure}[t]
\begin{center}
\includegraphics[width=0.7\textwidth]{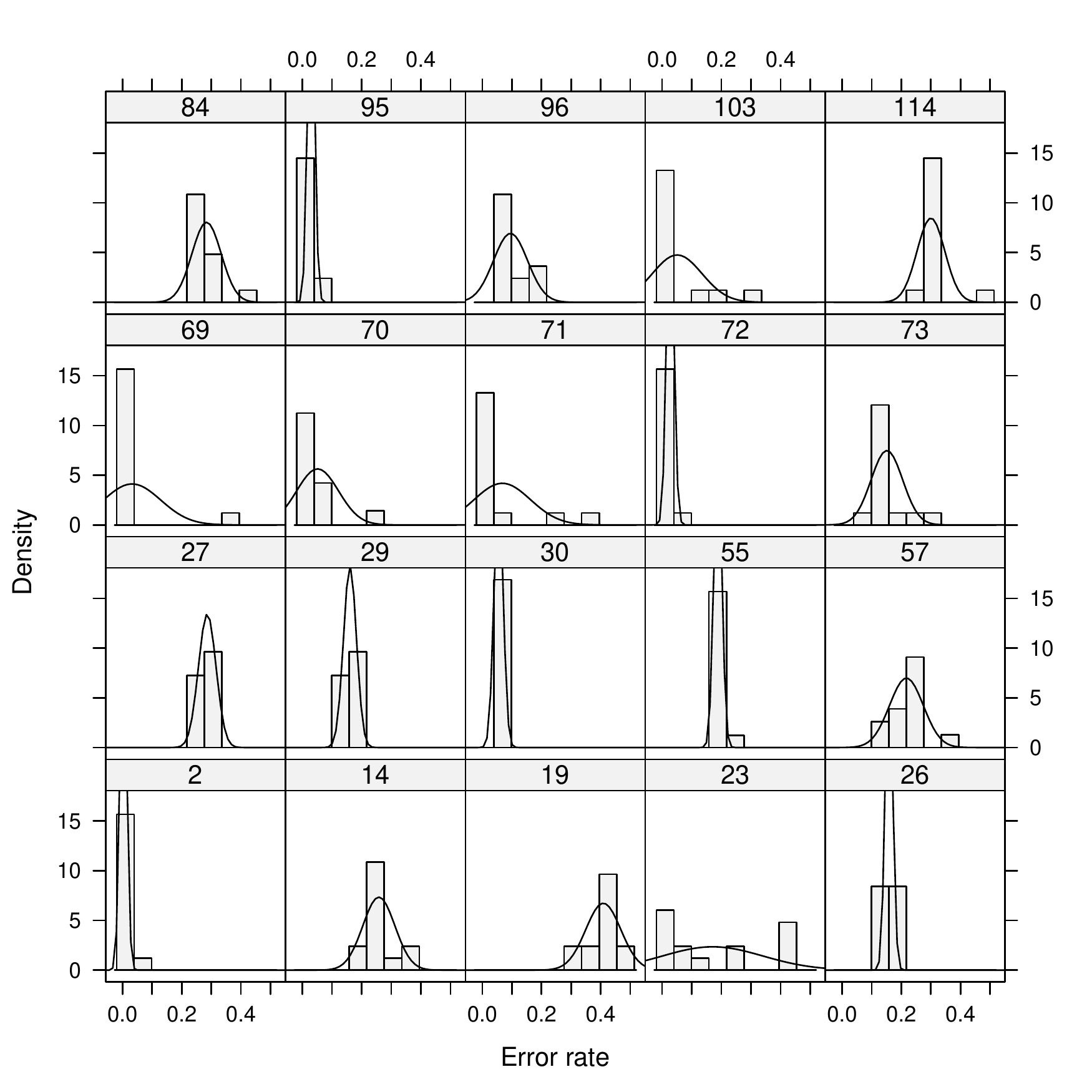}
\caption{Histogram of error rates for 20 random datasets across the
  different algorithms, and the best fit Gaussian distribution.
\label{hist2}}
\end{center}
\end{figure}

Figure~\ref{hist1} plots the histogram of the error rates across the
115 datasets for the different algoritms, superimposed with the best
fit Gaussian. One can argue that the distribution of error rates can
reasonably be considered as normal, which would match the assumptions
of the equation~\ref{eq2}. Similarly, Figure~\ref{hist2} is the
distribution of error rates for 20 random datasets, and again one
argue that the assumption in equation~\ref{eq3} is reasonable.

A more formal model verification is the procedure of posterior
predictive check proposed by \cite{gelman1996posterior}. The
procedure calculates how unlikely is the true data when compared to
data generated by the model, given the posterior distribution of the
hyperparameters of the model. The data (true and generated) are
summarized by the $\chi^2$ discrepancy \citep{gelman1996posterior}.

\begin{figure}[t]
\begin{center}
\includegraphics[width=0.7\textwidth]{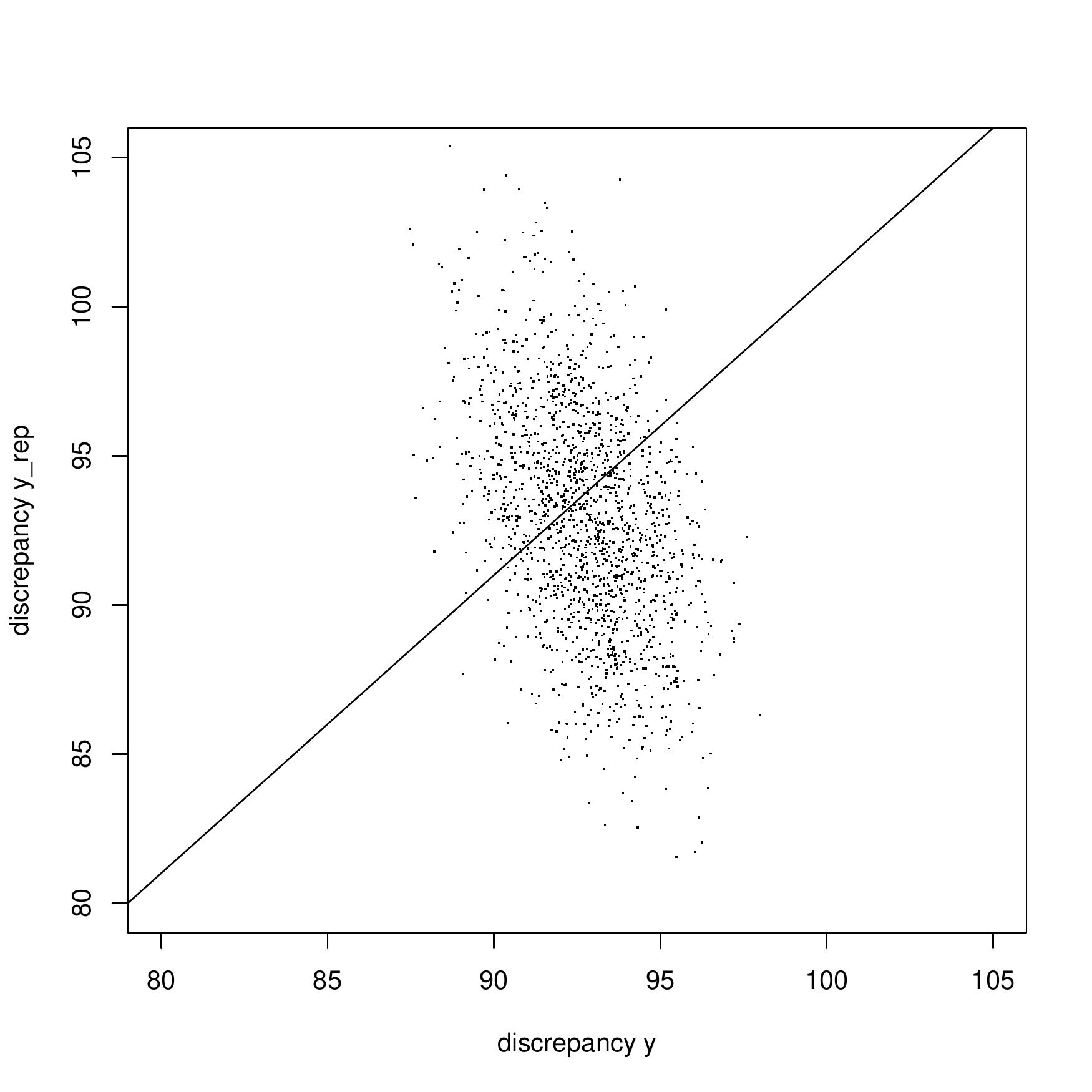}
\caption{Discrepancy ($\chi^2$) of the generated data against the
  discrepancy of the true data, for 1667 data from the MCMC chains
\label{discr}}
\end{center}
\end{figure}

Figure~\ref{discr} shows the relation between the discrepancies of the
true data and that of the generated data, for 1667 of the points
generated by the MCMC algorithm, following the posterior of the
parameters of the model. The probability of that the real data (or
data with even more extreme discrepancies) were generated by the model
is the proportion of the points above the $x=y$ line. As one can see,
that probability is around 0.5, and the real data is clearly among the
most likely of the data generated by the model. Thus, under this
criterion, the model is appropriate. But we must also point out that
3\% of the data generated was negative, which is at least esthetically
unpleasant.

\section{Convergence of the MCMC}
\label{bayes2}

There was no attempts to optimize the number of simulations of the
MCMC algorithm. We used JAGS, with 5000 burnin steps and 5000
adaptative steps. Finally we ran 100000 total interactions on 4
separated chains. 

Below is the Gelman and Rubin diagnostic \citep{brooks1998general}
which compares the variance within and between chains, as reported by the
function \texttt{gelman.diag} from the R package \texttt{coda}, where the
variables saved are the ones in Equation~\ref{eq8}: \texttt{b0} is
$\alpha$ and \texttt{b1[a]} are the $\beta_a$ for each algorithm, \texttt{b2[d]} are the $\delta_d$ for each dataset and
\texttt{ySigma} is $\sigma_0 $ from Equation~\ref{eq1}, \texttt{a1SD} and
\texttt{a2SD} are $\sigma_a$ and $\sigma_d$ from \ref{eqz4} and
\ref{eqz5}. Not all lines that refer to the \texttt{b2[d]} variables
are shown, but they all have the same values.

\begin{verbatim}
Potential scale reduction factors:

        Point est. Upper C.I.
b0               1          1
b1[1]            1          1
b1[2]            1          1
b1[3]            1          1
b1[4]            1          1
b1[5]            1          1
b1[6]            1          1
b1[7]            1          1
b1[8]            1          1
b1[9]            1          1
b1[10]           1          1
b1[11]           1          1
b1[12]           1          1
b1[13]           1          1
b1[14]           1          1
b2[1]            1          1
b2[2]            1          1
b2[3]            1          1
...
b2[113]          1          1
b2[114]          1          1
b2[115]          1          1
ySigma           1          1
a1SD             1          1
a2SD             1          1

Multivariate psrf

1
\end{verbatim}
Values between 1 and 1.1 suggest convergence
of the interactions. 

The effective sizes of the interactions shows that there is no problem
of high autocorrelation. Again not all values for the \texttt{b2[d]}
variables are shown.
\begin{verbatim}
       b0     b1[1]     b1[2]     b1[3]     b1[4]     b1[5]     b1[6]     b1[7] 
100000.00 101017.13  98356.57  99052.19  99172.48  99820.25  99350.88 100262.05 
    b1[8]     b1[9]    b1[10]    b1[11]    b1[12]    b1[13]    b1[14]     b2[1] 
100541.69  98992.43  98427.81 100000.00  97006.48  97467.65  94816.02 100000.00 
    b2[2]     b2[3]     b2[4]     b2[5]     b2[6]     b2[7]     b2[8]     b2[9] 
 99043.72  97966.91  99380.08 100000.00  99530.18 101053.74 100087.01 100000.00 
 ...
  b2[114]   b2[115]    ySigma      a1SD      a2SD 
100204.87 100216.75  55353.48  16108.44  51977.11 
\end{verbatim}

% Not all tests generate all positive results, but none seems to show a
% systematic failure of convergence. For example
% the Heidelberger diagnostics \citep{heidelberger1981spectral} shows
% that one variable (\texttt{b1[3]}) failed the stationary test in one
% of the chains but not on the other two chains, while some other
% variable (\texttt{b1[8]}) failed the halfwidth test in another
% chain. 

\section{Results with the robust Bayesian model}
\label{app:robust}

This section displays the results of the Bayesian analysis based on
the robust model. Table~\ref{xtabbanova} is the full probability table
from the Bayesian ANOVA. 

% latex table generated in R 3.3.0 by xtable 1.8-2 package
% Wed Jun  1 17:55:55 2016
\begin{table}[ht]
\tiny
\centering
\begin{tabular}{rrrrrrrrrrrrrr}
  \hline
 & rf & svmRadial & gbm & nnet & rknn & svmPoly & knn & svmLinear & glmnet & elm & lvq & sda & nb \\ 
  \hline
svmRadial & 1.00 &  &  &  &  &  &  &  &  &  &  &  &  \\ 
  gbm & 1.00 & 1.00 &  &  &  &  &  &  &  &  &  &  &  \\ 
  nnet & 0.64 & 0.93 & 0.95 &  &  &  &  &  &  &  &  &  &  \\ 
  rknn & 0.74 & 0.96 & 0.97 & 1.00 &  &  &  &  &  &  &  &  &  \\ 
  svmPoly & 0.50 & 0.85 & 0.88 & 1.00 & 1.00 &  &  &  &  &  &  &  &  \\ 
  knn & 0.45 & 0.83 & 0.87 & 1.00 & 1.00 & 1.00 &  &  &  &  &  &  &  \\ 
  svmLinear & 0.07 & 0.32 & 0.38 & 0.99 & 0.98 & 0.99 & 1.00 &  &  &  &  &  &  \\ 
  glmnet & 0.12 & 0.43 & 0.49 & 0.99 & 0.98 & 1.00 & 1.00 & 1.00 &  &  &  &  &  \\ 
  elm & 0.00 & 0.01 & 0.01 & 0.61 & 0.49 & 0.73 & 0.78 & 0.98 & 0.96 &  &  &  &  \\ 
  lvq & 0.00 & 0.00 & 0.01 & 0.53 & 0.41 & 0.66 & 0.71 & 0.97 & 0.94 & 1.00 &  &  &  \\ 
  sda & 0.00 & 0.01 & 0.01 & 0.54 & 0.42 & 0.66 & 0.71 & 0.97 & 0.94 & 1.00 & 1.00 &  &  \\ 
  nb & 0.00 & 0.00 & 0.00 & 0.00 & 0.00 & 0.00 & 0.00 & 0.05 & 0.03 & 0.39 & 0.45 & 0.44 &  \\ 
  bst & 0.00 & 0.00 & 0.00 & 0.00 & 0.00 & 0.00 & 0.00 & 0.00 & 0.00 & 0.01 & 0.01 & 0.01 & 0.52 \\ 
   \hline
\end{tabular}
\caption{The probability that the difference between the error rate is
  within the limits of irrelevance (from -0.0112 to 0.0112) for all
  pairs of algorithms using the robust model.}\label{xtabbanova}
\end{table}

We cannot perform the verification of the model using posterior
predictive check because the $\chi^2$ discrepancy needs the variance
of the data. Under the robust model, the variance of the data depends
also on the degrees of freedom of the student-t distribution, and in
the case of the robust simulations is 1.12, and the variance of the student-t
distribution is not defined for degrees of freedom below 2.

\end{document}